\documentclass[acmlarge]{acmart}
 \pdfoutput=1
\AtBeginDocument{%
  \providecommand\BibTeX{{%
    \normalfont B\kern-0.5em{\scshape i\kern-0.25em b}\kern-0.8em\TeX}}}

\setcopyright{acmcopyright}
\copyrightyear{2020}
\acmYear{2020}
\acmDOI{00.0000/0000000.0000000}

\acmJournal{IMWUT}
\acmVolume{4}
\acmNumber{3}
\acmArticle{100}
\acmMonth{0}

\usepackage{graphicx}
\usepackage{subfigure} 
\usepackage{pifont}
\usepackage{amssymb}
\newcommand{\cmark}{\ding{51}}%
\newcommand{\xmark}{\ding{55}}%
\usepackage{multirow}
\usepackage{diagbox}
\usepackage{array}
\usepackage{float}
\usepackage{soul}
\usepackage[ruled, linesnumbered]{algorithm2e}
\usepackage{tablefootnote}
\usepackage{makecell}
\usepackage{amsmath}
\DeclareMathOperator*{\maximize}{MAX}

\begin{document}

\title{ESPRESSO: Entropy and ShaPe awaRe timE-Series SegmentatiOn for processing heterogeneous sensor data}

\author{Shohreh Deldari}
\email{shohreh.deldari@rmit.edu.au}
\orcid{0002-6958-3086}
\affiliation{%
  \institution{School of Science, RMIT University/ Data61, CSIRO}
  \city{Melbourne}
  \state{VIC}
  \country{Australia}
  \postcode{3000}
}

\author{Daniel V. Smith}
\email{daniel.v.smith@data61.csiro.au}
\orcid{0001-9983-095X}
\affiliation{%
  \institution{Data61, CSIRO}
  \city{Hobart}
  \state{TAS}
  \country{Australia}
}
\author{Amin Sadri}
\email{amin.sadri@anz.com}
\affiliation{%
  \institution{ANZ}
  \city{Melbourne}
  \state{VIC}
  \country{Australia}
  \postcode{3000}
}
\author{Flora Salim}
\email{flora.salim@rmit.edu.au}
\orcid{xxxx-xxxx-xxxx}
\affiliation{%
  \institution{School of Science, RMIT University}
  \city{Melbourne}
  \state{VIC}
  \country{Australia}
  \postcode{3000}
}

\renewcommand{\shortauthors}{Shohreh Deldari, Daniel V. Smith, Amin Sadri, Flora D. Salim}

\begin{abstract}

Extracting informative and meaningful temporal segments from high-dimensional wearable sensor data, smart devices, or IoT data is a vital preprocessing step in applications such as Human Activity Recognition (HAR), trajectory prediction, gesture recognition, and lifelogging. In this paper, we propose \textit{ESPRESSO} (Entropy and ShaPe awaRe timE-Series SegmentatiOn), a hybrid segmentation model for multi-dimensional time-series that is formulated to exploit the entropy and temporal shape properties of time-series. \textit{ESPRESSO} differs from existing methods that focus upon particular statistical or temporal properties of time-series exclusively. As part of model development, a novel temporal representation of time-series $WCAC$ was introduced along with a greedy search approach that estimate segments based upon the entropy metric. \textit{ESPRESSO} was shown to offer superior performance to four state-of-the-art methods across seven public datasets of wearable and wear-free sensing. In addition, we undertake a deeper investigation of these datasets to understand how ESPRESSO and its constituent methods perform with respect to different dataset characteristics. Finally, we provide two interesting case-studies to show how applying \textit{ESPRESSO} can assist in inferring daily activity routines and the emotional state of humans.

%
\end{abstract}



\maketitle
\section{Introduction}

Today, there is a growing demand for data mining technologies to transform the complex, unwieldy data collected from a broad diverse range of wearable devices, smartphones, and sensors into compact and actionable information. Whilst supervised methods work well, they require carefully labelled samples. Annotating datasets of wearable sensors can be challenging for a couple of reasons. In addition to the privacy issues associated with collecting human data, the huge volume of data and hierarchical structure of human activities can make the annotation process time-consuming, expensive and sometimes even infeasible. Consequently, unsupervised and self-supervised techniques have gained a lot of attention \cite{saeed2019multi}. Furthermore, automatic knowledge extraction techniques are required to factorise this large volume of sensor data into interpretable pieces of information. \par

Time-series segmentation is the process of partitioning time-series into a sequence of discrete and homogeneous segments. We propose a new multivariate time-series segmentation technique 
to be used as a preliminary processing or exploratory data analysis step prior to 
tasks such as prediction, feature selection, semi-supervised or unsupervised classification. 
Therefore, the motivation of this paper is to enable a deeper unsupervised exploration of wearable sensor datasets by factorising them into a set of atomic primitives of physical action or emotion. By enabling these primitives to be discovered, the process of feature engineering can be accelerated, whilst in addition, greater insights into the underlying properties of the data can be learned. 
\par

Recent studies in Human Activity Recognition (HAR) have demonstrated the effectiveness of using temporal segmentation in combination with classification \cite{aminikhanghahi2019enhancing,liono2016optimal,shoaib2016complex, wang2018modeling, chamroukhi2013joint}. 
In addition to HAR, time-series segmentation have been applied to other modeling tasks with wearable sensors, including trajectory prediction \cite{Sadri2018tajectory}, motion-based user authentication \cite{huang2019id}, life-logging \cite{chavarriaga2013opportunity}, elderly rehabilitation \cite{lam2016automated}, 
anomaly detection techniques \cite{RAJAGOPALAN20063309}, predictions \cite{song2018evolutionary,Aminikhanghahi2017,song2016multivariate}, and feature selection \cite{lee2018time}.

\noindent
\begin{figure}[b]
\centering
\subfigure[][]{%
\label{fig:stat}%
\includegraphics[width=.49\linewidth, height = 1in]{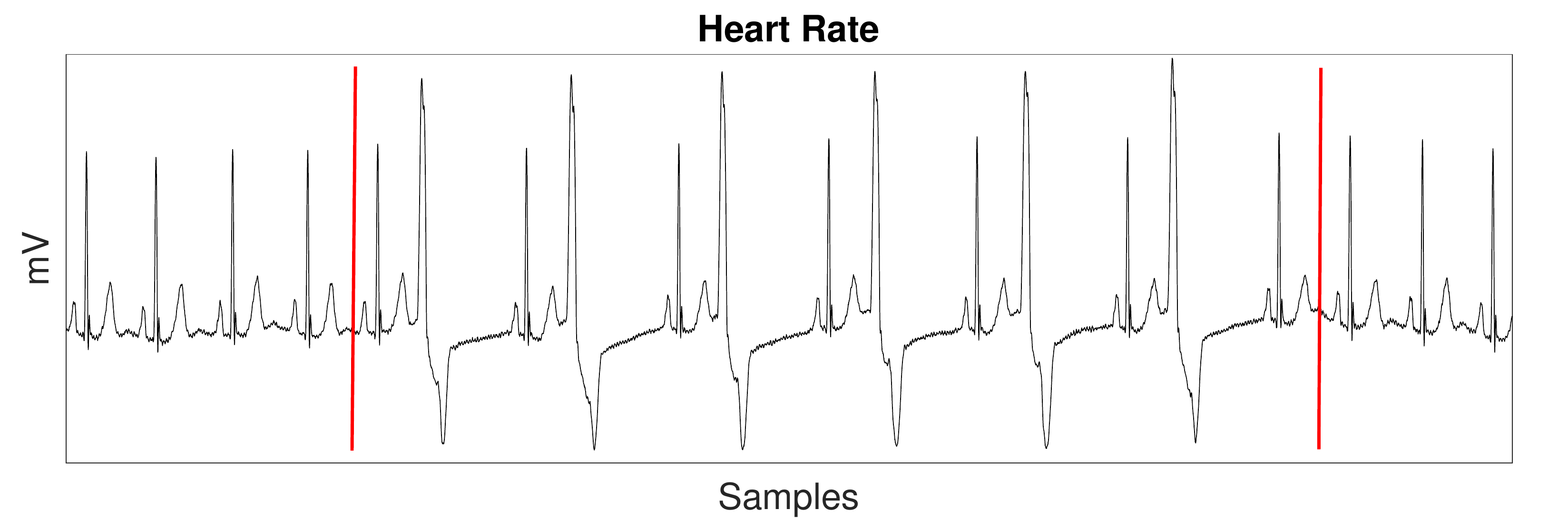}}%
\hspace{0.1pt}%
\subfigure[][]{%
\label{fig:heart}%
\includegraphics[width=.49\linewidth, height= 1in]{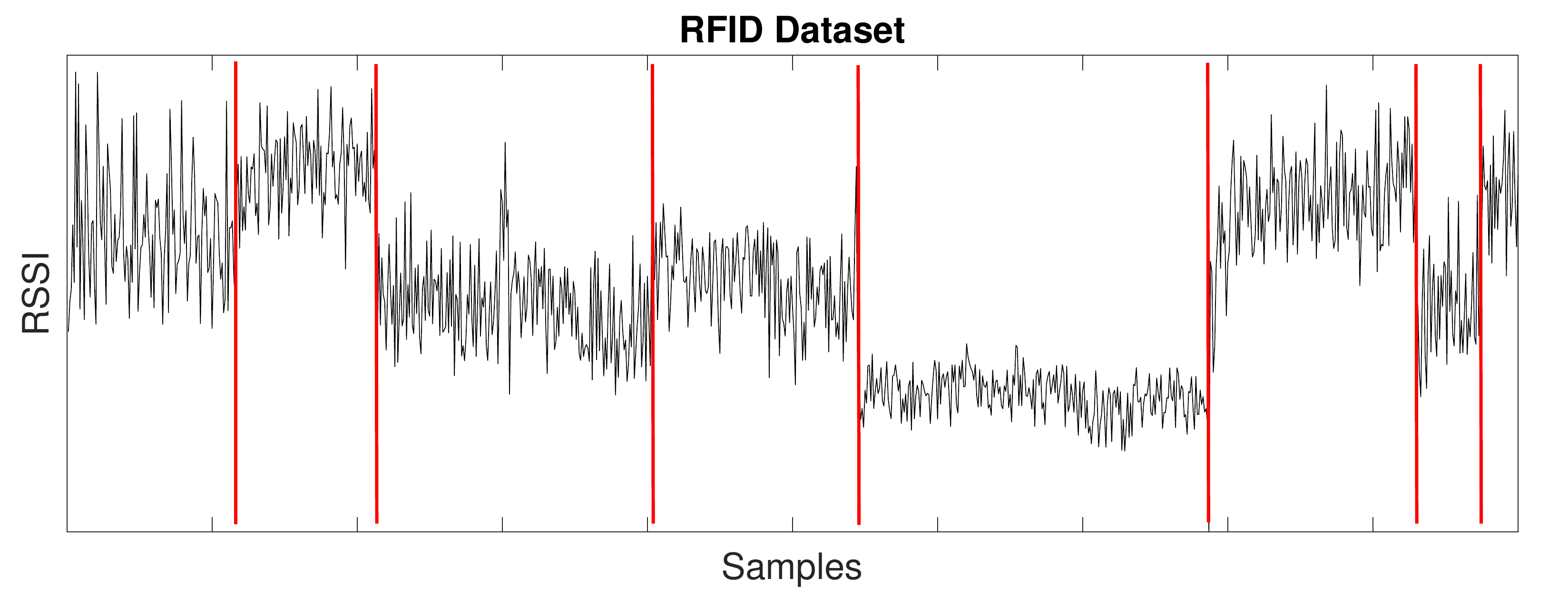}}%

\caption{a) A time-series of human heart beats with segments that have clear differences in the temporal shape. b) A time-series of RSSI measured by a RFID tag where segments of unique human postures possess clear statistical differences.}
\label{fig:example1}
\end{figure}

In pervasive computing applications, the time-series being collected will often be heterogenous encompassing a diverse range of characteristics with respect to their dimensionality, continuity, statistical properties and shape. Figure \ref{fig:example1} shows two time-series with very different properties. Figure \ref{fig:example1}(a) shows the repetitive temporal shape patterns of the human heart measured with a wearable electrocardiogram (ECG). Figure \ref{fig:example1}(b) shows a sequence of human postures that have been measured with a passive RFID tag array; the RSSI of each posture have different statistical properties. This figure clearly shows the semantics of each use case should be extracted by exploiting different time-series properties. Statistical changes can be used to segment the human postures with high precision, however, temporal shape changes will fail in distinguishing these segments. In contrast, exploiting temporal shape changes in the ECG data will be advantageous to segment abnormal heartbeats (the middle segment of the ECG) compared to using statistical changes that are more uniform across the segments in (a) than (b). 

While current time-series segmentation methods exploit individual characteristics of the signal, that include the temporal shape, statistics, or probability distribution, we propose Entropy and ShaPe awaRe time-series SEgmentation (\textit{ESPRESSO}), a hybrid model that incorporates multiple signal characteristics. 
 To achieve this, ESPRESSO integrates the search and score based mechanisms of segmentation through a newly proposed shape representation, \textit{WCAC}, and a greedy search that exploits a non-parametric entropy based cost function. The segmentation results are then further enhanced by devising an embedded channel ranking algorithm. \textit{ESPRESSO} has been developed to accurately segment a wider range of time-series by relaxing some of the assumptions imposed by statistical or temporal shape-based methods.


The main challenges of current multi-dimensional time-series segmentation approaches are as follows:
1) Model assumptions: Models make parametric assumptions about the underlying properties of time-series that can limit its application. 2) Model parameterisation: Segmentation models generally utilise a number of parameters and thresholds that need to be carefully tuned based upon domain knowledge. 3) Channel ranking: In multi-dimensional time-series, not all dimensions (channels) are equally important to achieving accurate segmentation. Although there are numerous supervised channel selection techniques for classification tasks, there is little work on ranking the relevance of channels for unsupervised segmentation.



Current work on temporal shape-based segmentation \cite{gharghabi2018domain} operates under the principle that similar shaped patterns are associated with the same segment class and occur within close temporal proximity. This assumption, however, can lead to degraded segmentation performance under any of the following conditions: 
 a) Several instances of the same class (with the same label) repeat multiple times across the time-series; b) segment classes are not comprised of repeated shape patterns; c) shape patterns drift with time. 
Each of these conditions are commonly encountered in wearable sensor use-cases. We propose a temporal shape-based segmentation method, \textit{Weighted Chain Arc Curve (WCAC)}, to address these limitations.
In addition to temporal shape-based methods, there are a range of statistical based segmentation approaches. Such approaches have commonly employed parametric models in the form of Probability Density Functions \textit{(PDF)} \cite{Basseville93,hallac2018greedy,Ni2016} and auto-regressive models \cite{Takeuchi2002} but have somewhat limited application given they impose strong assumptions upon the statistical properties of the time-series. 
Non-parametric kernel based methods have been proposed \cite{aminikhanghahi2019enhancing,Kawahara2009,liu2016complex,Yamadaahsic} to offer greater modelling flexibility, but can be difficult to train and provide poor estimates across smaller sample sets. 
\par

\subsection{Contributions}
The main contributions of this paper are as follows:
\begin{enumerate}

    \item  \textit{ESPRESSO} is a novel time-series segmentation approach which integrates temporal shape and entropy based properties of multidimensional time-series. Unlike most state of the art methods that require several carefully tuned parameters, \textit{ESPRESSO} only depends on one parameter which can be selected with minimal risk, given \textit{ESPRESSO}'s performance is shown to be relatively consistent with respect to this parameter. \textit{ESPRESSO} is shown to outperform four state-of-the-art segmentation methods in terms of their F-score and RMSE measure across seven public datasets of wearable sensors in this experiment.

    \item We propose \textit{WCAC} to address particular limitations of existing shape-based segmentation approaches. In contrast to other temporal shape methods, \textit{WCAC} can accommodate both repeated segments and shifts in temporal shape across the time-series.  
           
    \item An embedded channel ranking has been utilised in \textit{ESPRESSO} to make segmentation more robust to noisy and/or irrelevant channels.
    
   
   \item We categorise time-series datasets with regards to their continuity and repetitive patterns. An ablation study is performed to evaluate \textit{ESPRESSO}'s performance with respect to these categories.
   
   \item Finally, we demonstrate the interpretability of segmentation results obtained by \textit{ESPRESSO}  for two real-world use-cases. The first study shows the ability of \textit{ESPRESSO} to discover deviations in the daily routines of people through life-logging data. The second study shows \textit{ESPRESSO} can identify the emotional states of people and provide an interpretation of their emotional transitions.
 \end{enumerate}   

\par



\section{Related works}


Although the term segmentation is frequently used in the time-series literature, we focus upon approaches that partition time-series from the bottom up by using salient changes in the series to identify individual segment boundaries, or from the top down by identifying the segment boundaries that optimise a cost function across an entire time-series. First, we review current applications of time-series segmentation in the field of wearable sensors and device-free dataset. We then provide an overview of general time-series segmentation approaches and  highlight the current limitations of these approaches.

\subsection{Wearable Sensors Segmentation}
For wearable sensors, segmentation methods use a fixed-length sliding window.. Authors in \cite{shoaib2016complex} have compared the effect of window size in detecting different types of activities in HAR applications. To estimate the effect of window size on activity recognition, they divide activities into two main groups: simple activities with periodic actions such as running and waving and more complex, non-periodic actions such as drinking coffee. They found that shorter windows were shown to be effective for simple periodic activities but less reliable to represent the more complex activities. Consequently, recent works, such as \cite{Peng2018Aroma}, have exploited different sized sizes to represent activities of varying complexity. The \cite{liono2016optimal} method proposed an optimization approach to find the optimal window size for activity segmentation, however, this can be challenging task to undertake given the variety of sensors and activities associated with real-world applications. \par

 Temporal segmentation  has been shown to be an important pre-processing step in high-dimensional wearable and device-free sensor applications \cite{lin2016movement, haladjian2019wearables, bulling2014tutorial}. 
Segmentation has been mentioned as an open challenge in analyzing life-logging data from wearable sensors in order to have more accurate models \cite{chavarriaga2013opportunity}. A recently published wearable development toolkit, \textit{WDK}, has incorporated time-series segmentation methods; this demonstrates the impact that these technique have had on wearable sensing applications \cite{haladjian2019wearables}. 
Authors in \cite{aminikhanghahi2019enhancing, aminikhanghahi2017using} showed that applying classification on top of a segmentation method produced more accurate results than performing classification with a fixed-length sliding window. They proposed a new probability metric, \textit{SEP} to improve current probability density-ratio change point detection techniques in smart home applications. The authors in \cite{wang2018modeling} propose a simple segmentation method that exploit knowledge of the statistical characteristics of low-level human activities (i.e. walking, running) to segment RFID signals. Temporal segmentation of motion data has been studied extensively. The method in \cite{noor2017adaptive} proposed using decision trees to find the split points. The authors in \cite{chamroukhi2013joint} proposed a multiple regression model based upon \textit{Expectation Maximization}, \textit{MRHLP}, which identifies activity boundaries as the points where there is a switch in the underlying models.

In addition to classification problems, the authors in \cite{Sadri2018tajectory,SadriSRSKM18} have shown that using temporal segmentation inconjunction with a prediction model leads to performance improvements. They have utilized an entropy-based temporal segmentation method to improve the prediction quality of user activity trajectories. For a user identification and authentication application, \cite{Huang2019AuId} devised a sequence labelling based segmentation approach to extract physical and behavioural characteristics of individuals from a sequence of their daily activities. Segmentation has also been used for feature selection in datasets of multi-dimensional human activity motion, electroencephalogram (EEG) signals and speech signals \cite{lee2018time}. \cite{sarker2017individualized, sarker2018mining} found temporal segmentation could be used to identify a user's behavioral characteristics from their smartphone usage data.
Table \ref{tab:relatedwork} summarizes some recent wearable sensor applications that benefit from unsupervised segmentation.
Other than human-centric applications, time-series segmentation has been applied to a broad range of fields such as sensor data processing, environmental modelling, financial events, music and speech processing, energy consumption predictions and so on. \cite{Aminikhanghahi2017} provides a detailed review of time-series change point detection methods.
\par

\begin{table}[]

\caption{List of some of  recent wearable sensor analysis applications which benefit from automatic segmentation.}
\label{tab:relatedwork}
\begin{tabular}{c|c|l}
\hline
\textbf{Year}         & \textbf{Paper}                                      & \textbf{Applying segmentation for ...}                                                                                     \\ \hline
\multirow{3}{*}{2019} & \cite{aminikhanghahi2019enhancing} & Activity recognition in smart-home                                                                       \\
                      & \cite{Huang2019AuId}               & Authentication and Identification                                                                        \\
                      & \cite{lee2018time}                 & \begin{tabular}[c]{@{}l@{}}Feature selection in HAR and EEG data\end{tabular} \\ \hline
\multirow{2}{*}{2018} & \cite{Sadri2018tajectory}          & User trajectory prediction                                                                               \\
                      & \cite{wang2018modeling}            & HAR using reflection of RFID signals                                                                     \\ \hline
\multirow{2}{*}{2017} & \cite{sarker2017individualized}    & \begin{tabular}[c]{@{}l@{}}User behavioral characteristic base on smartphone usage\end{tabular}      \\
                      & \cite{noor2017adaptive}            & \begin{tabular}[c]{@{}l@{}}Physical activity recognition using ACC data\end{tabular}                 \\ \hline
\multirow{2}{*}{2016} & \cite{liu2016complex}    & \begin{tabular}[c]{@{}l@{}} Complex activity recognition\end{tabular}      \\
                      & \cite{lin2016movement}            & \begin{tabular}[c]{@{}l@{}}Human motion modelling using ACC data\end{tabular}                 \\ 
                       & \cite{lam2016automated}            & \begin{tabular}[c]{@{}l@{}}Exercise and physical rehabilitation analysis\end{tabular}                 \\ 
                      \hline
\multirow{2}{*}{2013} & \cite{chamroukhi2013joint}    & \begin{tabular}[c]{@{}l@{}} Physical activity recognition using ACC data\end{tabular}      \\
                      & \cite{li2013daily}            & \begin{tabular}[c]{@{}l@{}}Exercise and physical rehabilitation for elderly people\end{tabular}               
                       \\ 
                        & \cite{lin2013online}            & \begin{tabular}[c]{@{}l@{}}Exercise and physical rehabilitation analysis\end{tabular}               
                       \\\hline
\end{tabular}
\end{table}

\subsection{Time-series Segmentation Approaches}

Time-series segmentation approaches can be divided into supervised and unsupervised techniques. Supervised methods infer the class labels of underlying time-series using binary or multi-class classifiers formed from Hidden Markov Models \cite{san2016segmenting} and Decision Trees \cite{noor2017adaptive} in order to identify segment boundaries. 
Unsupervised techniques are more commonly utilized than supervised approaches given they do not require training sets of segmented data. Instead unsupervised methods exploit the underlying signal properties to estimate the change points.
\subsubsection{Statistical approaches}

The statistical properties of a time-series are most frequently exploited in unsupervised segmentation. These methods can be categorised as top-down optimisation approaches \cite{hallac2018greedy,sadri2017information} that search for the set of segments that maximise its particular cost function, or bottom up approaches that identify individual segment boundaries from local deviations in the time series \cite{Basseville93,Kawahara2009,liu2013change,Ni2016,Takeuchi2002,Yamadaahsic}. 

Sadri proposed a top-down temporal segmentation method, \textit{IGTS}, which was based upon the information gain (IG) metric \cite{sadri2017information}. Segment boundaries were estimated by using a dynamic programming approach to maximise the IG of its constituent segments. A similar top-down approach was used in \cite{hallac2018greedy}, where a greedy search was used to identify the segment boundaries that maximize the regularized likelihood estimate of a segmented Gaussian model.

The statistical differences between time intervals have commonly been measured with the likelihood ratio formulation \cite{Basseville93,Kawahara2009,liu2013change,Ni2016,Takeuchi2002}. Within this formulation, parametric models have been used to estimate the intervals as Probability Density Functions (PDFs) \cite{Basseville93,Ni2016}, auto-regressive models \cite{Takeuchi2002} or state space models \cite{Kawahara2007}.
The parametric assumptions of these models, however, limit the types of statistical changes that can be detected. For instance, by fitting
a Gaussian distribution to segments such as in \cite{Basseville93,Ni2016}, only differences in the mean and/or standard deviation of adjacent intervals can be used in segmentation. Whilst these limitations can be relaxed by considering non-parametric density estimation, this still remains a difficult estimation problem to address.

Flexible non-parametric solutions \cite{Kawahara2009,liu2013change} were proposed to compute the likelihood ratio without the 
need for density estimation. It was found that estimating the ratio of PDFs directly was a simpler problem to address than density estimation. Hence, a non-parametric Gaussian kernel could be successfully used for this purpose. Kullback-Leibler Importance Estimation Procedure (KLIEP) was used to directly estimate the ratio of PDFs \cite{Kawahara2009}.
Liu adopted the Relative unconstrained Least Square Importance Fitting (\textit{RulSIF}) to directly estimate the relative ratio of PDFs \cite{liu2013change}. 
These non-parametric approaches for direct ratio estimation were challenging to train and required a cross-validation procedure for model selection. They also tend to produce poor estimates with small datasets. 
Yamada utilised the non-parametric \textit{additive Hilbert-Schmidt Independence Criterion(aHSIC)} for time-series segmentation \cite{Yamadaahsic}. Change points were detected by using the \textit{aHSIC} criteria to compute the dependency between time adjacent intervals and the pseudo label of statistical change between the intervals. 

Whilst non-parametric approaches offer greater flexibility to modeling statistical change
than the earlier parametric methods, they are not universally applicable to HAR applications. They assume statistical homogeneity within each segment and statistical heterogeneity between different segments. Whilst this assumption is appropriate for low level segmentation tasks, it will not always be valid for wearable sensing applications where extracted segments need to characterise complex actions, emotions and behaviours.



\subsubsection{Temporal shape approaches}

The temporal shape is another unique property of time-series that can be exploited in segmentation \cite{gharghabi2018domain,huang2014detecting} where changes in the temporal shape patterns of a time-series were used to estimate the segment boundaries. \textit{FLOSS}, Fast Low-Cost Semantic Segmentation \cite{gharghabi2018domain} works under the principle that patterns of similar shape were each associated with the same segment class and occur within close temporal proximity of each other. The limitations of such assumptions were described in the Introduction section. In contrast to \textit{FLOSS}, which is based on the most similar repeated patterns, the authors in \cite{huang2014detecting} proposed a segmentation model based on rare temporal patterns.
Although shape-based methods can be beneficial for time-series composed of repeated shape patterns, performance will degrade when segments are composed of diverse shapes or when the shape patterns of a segment drift over time. Recently, the authors of \cite{wang2019discovering,zhu2017chain} proposed a new pattern-based primitive, \textit{Chain}, to discover a chain of similarly shaped patterns. To make shape extraction robust against pattern drift, we customize this idea in our proposed shape-based segmentation method.

\section{Problem Definition}\label{problem_def}

While we follow the Matrix Profile framework introduced in \cite{yeh2016matrix}, for completeness, we firstly provide a definition of multi-dimensional time-series. \par

\textbf{Definition 1.} The high-dimensional time-series $X$, is an $D\times N$ matrix of $N$ samples and $D$ channels (time-series), such that $X=\{{X_{i}^{j} | 1<i<N, 1<j<D}\}$, where $X_i^j$ denotes the $i$th sample of the $j$th time-series channel.\par

\textbf{Definition 2.} The subsequence of a time-series, $X_{i,L}^{j}$, is a vector of samples in channel $j$ ranging between index $i$ and index $i+L-1$. $L$ is the length of the subsequence.\par

\textbf{Definition 3.} \textit{Matrix Profile}, $MP$, is a $D \times N-L+1$ matrix where $MP_{i}^{j}$ denotes the distance between subsequence $X_{i,L}^{j}$ and its nearest neighbor in $X^j$.
Here we employed Euclidean distance as a similarity metric to compare subsequences.\par
\textbf{Definition 4.} \textit{Matrix Profile Index} or $MPI$ is a $n \times n-L+1$ matrix where $MPI_i^j$ denotes the index of the nearest neighbor (the most similar subsequence) for the subsequence $X_{i,L}^{j}$.\par

According to this definition, the most similar pattern to subsequence $X_{i,L}^{j}$ is $MPI_i^j$ with the similarity distance of $MP_{i}^{j}$.
The authors in \cite{gharghabi2018domain} defined $Arc Curve$ on top of $Matrix Profile$ in their shape-based segmentation technique. \par
\textbf{Definition 5.} $Arc_i^j$ is an arc between $X_{i,L}^{j}$ and $MPI_i^j$ and $Arc Curve$, $AC$, is a vector of the same length of time-series $X^j$ where $AC_t^j$ denotes how many arcs, $Arc_i^j, 1 \leq i \leq N-L+1$, cross the $t$th time tick in $j$th channel.\par

\begin{figure*}[t]
    \centering
    \includegraphics[width=0.90\linewidth]{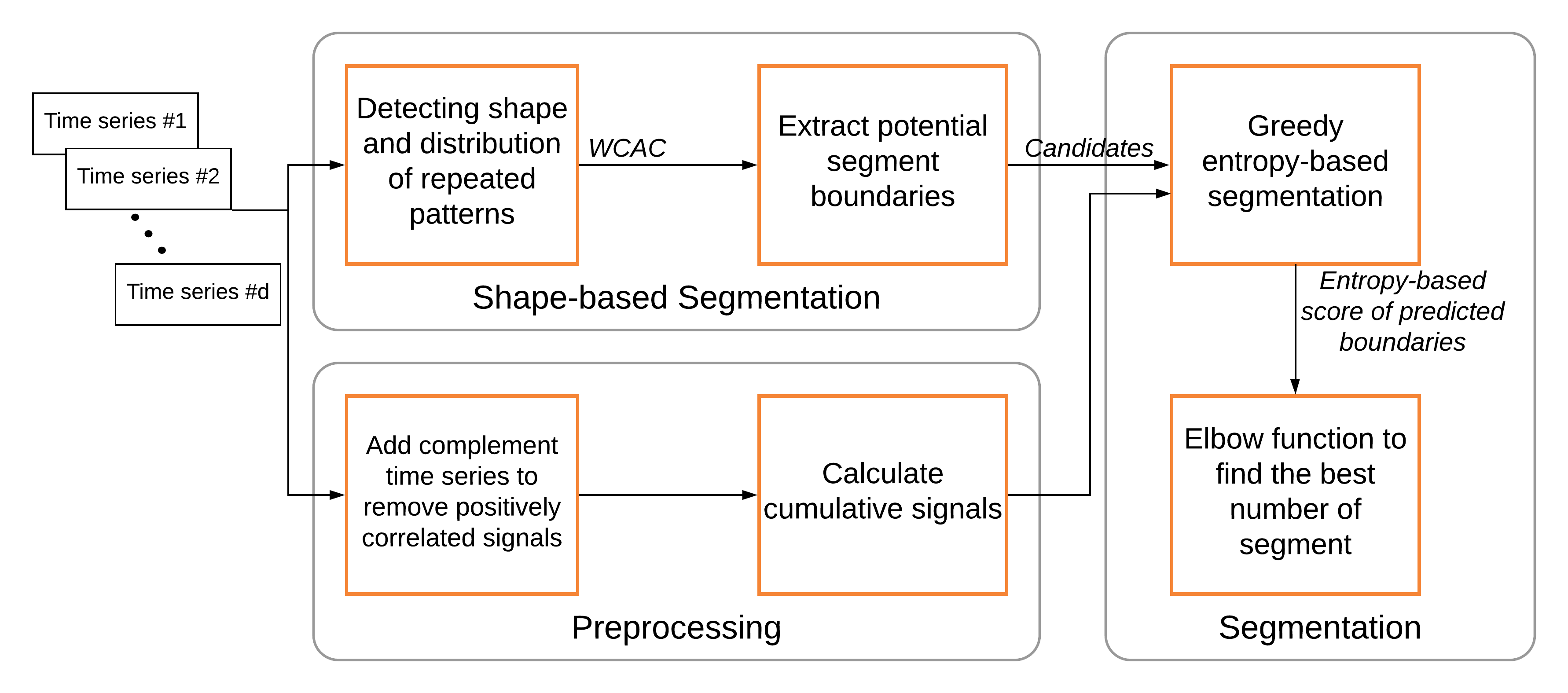}
    \caption{The workflow of the proposed time-series segmentation method(\textit{ESPRESSO}).}
    \label{fig:workflow}
\end{figure*}

\textbf{Problem definition.} \textit{Given the multi-dimensional dataset $X$ of $d$ time-series, we attempt to detect the $k-1$ transition times $t_1$,$t_2$,...,$t_{k-1}$ that are indicative of state changes in $X$. These transition times represent the boundaries needed to extract $k$ segments.}
Our proposed method consists of three steps: 

\begin{enumerate}
    \item  Extracting potential segment boundary candidates by analyzing the temporal shape across all dimensions;
    \item Employing a greedy search over boundary candidates in order to identify the set of segments with a minimum average entropy;
    \item Ranking channels and estimating the number of segments.
\end{enumerate}
Figure ~\ref{fig:workflow} provides a visual overview of the work flow associated with our method.\par

\section{ESPRESSO: Methodology} \label{ESPRESSO}
In this section we describe different parts of our proposed segmentation technique.

\subsection{Shape-based Segmentation}
The main assumption of our shape-based segmentation methods is that repeated patterns relate to the same class segments, and hence, occur within close temporal proximity. To extract the most similar shapes within each time-series, we utilized the \textit{MP} technique \cite{yeh2016matrix} and the $Arc$ definition \cite{gharghabi2018domain}. 
The FLOSS \cite{gharghabi2018domain} method works under the principle that the $AC$ will have its minimum values at segment boundaries based on the assumption that a large majority of arcs will be confined to individual segments with very few arcs crossing over segments. This assumption is more likely to be violated, however, if the dataset contains repeated segments. Figure 3 shows an example of a sequence of physical activities that starts with jumping, is followed by running and then returns to jumping again. In this case, running subsequences can find their most similar subsequence within the alternative running segment. This leads to a larger number of arcs spanning over the intermediate jumping segment, and hence, higher AC values being produced across this intermediate segment. This degrades the ability of FLOSS to estimate the activity transition times (comparing Figure 3(b) and Figure 3(c)). 
 In order to address this issue, FLOSS \cite{gharghabi2018domain} defined a temporal constraint parameter to ignore arcs that are longer than a threshold. Setting this threshold, however, requires having detailed knowledge about the particular problem in question. Furthermore, a threshold places an upper limit upon the segment size that can be considered; this is undesirable from an algorithm design perspective. 
 To address this problem, we propose a novel time-series representation primitive named \textit{Weighted Chained Arc Curve} (\textit{WCAC}) to capture the density of pattern repetition with time. The \textit{WCAC} evaluates the arc $Arc_{i}^j$ according to the temporal distance between each pair of similar subsequences. We show in Section \ref{sec:experiments} that we can achieve a far more accurate segmentation result by using \textit{WCAC} when compared to \textit{FLOSS}. 
\begin{figure}[]
\centering
\includegraphics[width=.5\linewidth]{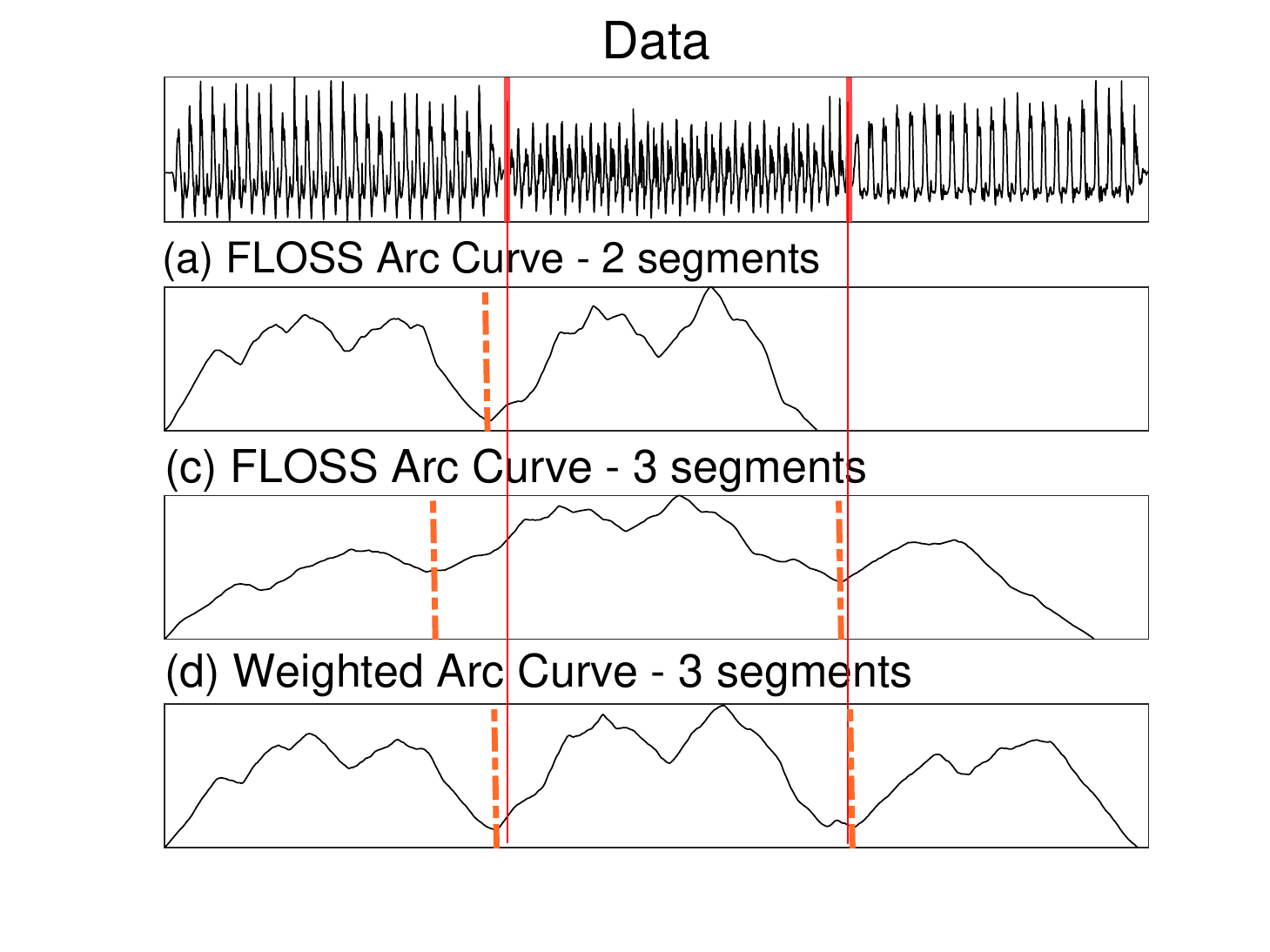}
  \caption{a)Accelerometer data for a sequence of [jumping,running,jumping]; b)\textit{FLOSS} over the first two segments; c)\textit{FLOSS} over all segments; d)WCAC over all segments.}
  \label{fig:three_segment_ac}
\end{figure}

 To define \textit{WCAC}, we first explain how a chain of similar arcs is calculated. To increase the robustness of the representation to noise and signal drift, we modified the \textit{AC} definition to consider a chain of similar subsequences instead of only the most similar subsequence. 
The authors in \cite{zhu2017chain} proposed the time-series \textit{Chain} as a new primitive to sit on top of the MP representation. We modified this chain definition to fit our problem. We believe that considering a chain of patterns is crucial in the context of Human Activity Recognition (HAR), given many activities are associated with motion patterns that can drift with time. We define $Chained Arc Curve$ as follows:\par

\textbf{Definition 6.} \textit{Chained Arc Curve}, $CAC_i^j =\{{x_{s1},x_{s2},...,x_{sl}}\},(1 \leq s1..sl \leq n-L+1) $, is a set of similar subsequences in the $j$th channel of input. $CAC_i^j$ is ordered in terms of their temporal distance to $X_i^j$, and for any $1 \leq si \leq l$, we have $ x_{si} = MPI_{x_{si-1}}^j $ and $x_{s1} = x_i $, where $n$ is the length of channel and $L$ is the size of the subsequence, $l$ denotes the length of the chain. \par

\begin{figure}[b]
\centering
\includegraphics[width=0.9\linewidth,height=1in]{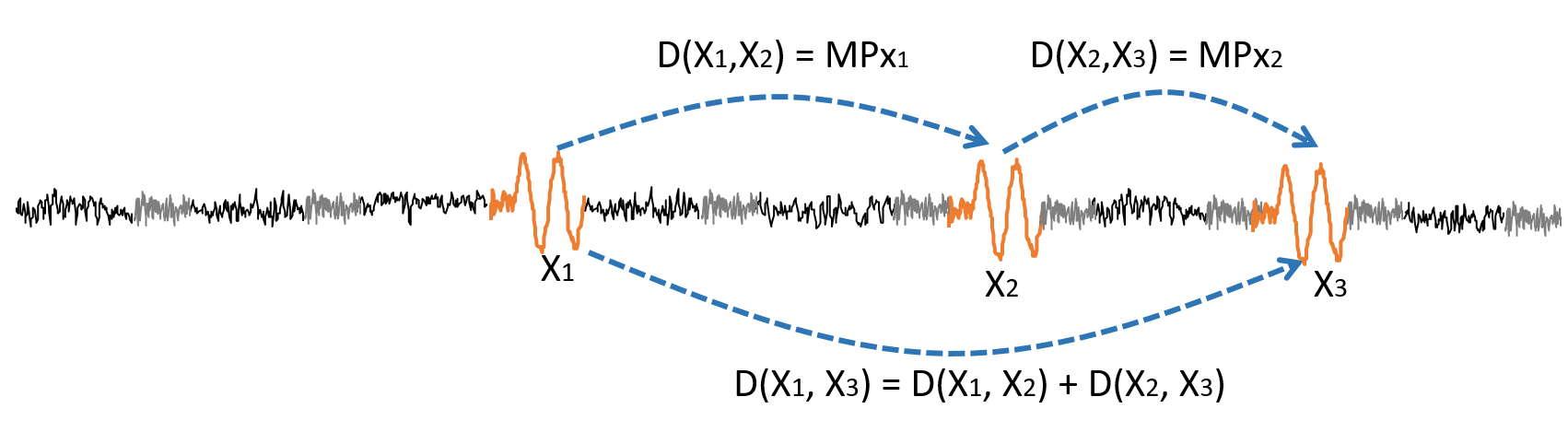}
\caption{Chained Arc Curve example.}
\label{fig:klevel}
\end{figure}

Figure \ref{fig:klevel} illustrates a simplified version of $CAC$ with arc chains of second order neighbours. If $X_2$ is the nearest neighbour (the most similar subsequence) of $X_1$, and $X_3$ is the nearest neighbour of $X_2$, then we add an arc between $X_1$ and $X_3$ (if there is no arc yet). The distance of the new arc will be considered as $D(X_1,X_2)+D(X_2,X_3)$. Any higher order arcs will be included in the $CAC$ if the distance is less than the specified threshold.
 To avoid trivial matches, we ignore similar subsequences in an exclusion zone of $L/2$ samples before and after the location of the query. In order to ensure the extracted patterns in the chain are similar, we limit the length of the chain by defining a distance threshold between the first and last subsequence of the chain. 

The other modification is to consider the locality of repeated patterns. Each arc in the $CAC$ is assigned a weight as an inverse function of its length. Consequently, arcs of a smaller length were provided with greater weight, given they were more likely to belong to the same segment instance than the arcs of a greater length, which were more likely to cross over other segments.
\textit{WCAC}, is defined as follows:
\begin{equation}
    WCAC_{i}^{j} = \sum_{A}^{} {\frac{MP_a^j}{|a - MPI_a^j|}}  , A=\{Arc_{a}^j \in CAC^j|  i \in [a,MPI_a^j]\}
\end{equation}
Figure 3(d) shows that $WCAC$ is a more effective representation to estimate change points, as unlike the original $AC$ (Figure 3(c)), there are two local minima in close proximity to the actual segment boundaries.\par

\noindent
  \begin{minipage}[t]{.5\textwidth}
  \raggedleft
 \begin{algorithm}[H]
\SetAlgoLined
\KwData{MPindex, MP}
\KwResult{Chain of extracted arcs for each subsequence(CAC), NewArcsSimilarity}

 $CAC$= MPindex\;
 $newArcs$ = MPindex(MPindex)\;
 \While{newArcs}{
  $NewArcsSim$ = MP + MP(newArcs)\;
  \For{each $arc$ in $newArcs$}{
  \eIf {$newArcsSim(arc)$ < threshold}
   { add $arc$ to $CAC$\;
   }{
    remove $arc$ from $newArcs$  \;
  }
}
  \textit{newArcs = MPindex(newArcs)}\;
 } 
 return $CAC$, $newArcsSim$\;
 \caption{ExtractCAC}
 \label{alg:cac}
\end{algorithm}
\end{minipage}%
\begin{minipage}[t]{.5\textwidth}
\raggedright
\begin{algorithm}[H]
\SetAlgoLined
\KwData{MP, MPindex}
\KwResult{Weighted repesentation of chain of arcs(WCAC)}
 $CAC$, $NewArcsSim$ =
       ExtractCAC($MP$,$MPindex$)\;
 \#initialize \;
 temporalDistance = normalize\_distance(CAC)\;
 $WCAC$ = zeros(length($MP$))\;
 \#calculate weight for each extracted arc in chains\;
 \For{each $arc$ in $CAC$}{
  $weights$ = $NewArcsSim(arc)$/temporalDistance($arc$)\;
  $WCAC$ = $WCAC$($arc_{start}$ - $arc_{end}$) + $weights$ \;
  }
  return $WCAC$\;
 \caption{ExtractWCAC }
 \label{alg:wcac}
\end{algorithm}
\end{minipage}

The algorithm to compute the $CAC$ and $WCAC$ are provided in Algorithm \ref{alg:cac} and \ref{alg:wcac} respectively. 
The chain ($CAC$) is initialized with the MP index that represents the arcs between each subsequence and its nearest neighbour subsequence, which is the most similar subsequence too (line 1). A set of second-level arcs are then constructed between each subsequence and the nearest neighbor of its own nearest neighbor (line 2). We then look for next level arcs to add to the current chain (lines 3-13). If the new arc meets the distance condition (line 6), it will be added to the chain in line 7. This process is repeated until there are no new arcs to add. To calculate $WCAC$ (Algorithm \ref{alg:wcac}), for each arc in $CAC$, the weight will be updated according to the similarity ($NewArcsSim$) and normalized arc length  ($temporalDistance$) in lines 6-9. \par

\begin{figure}[]
\centering
  \centering
  \includegraphics[width=.69\linewidth]{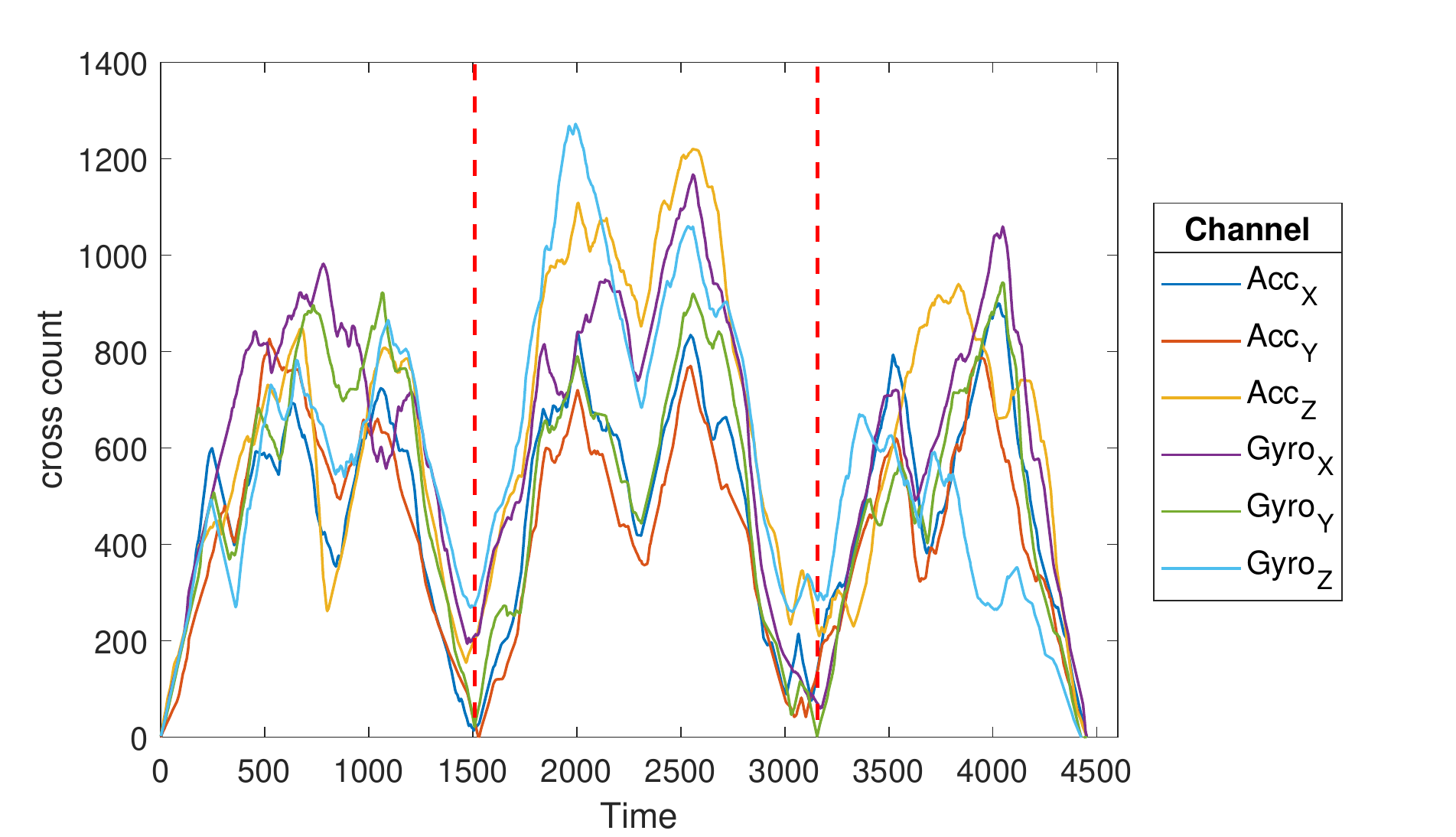}
  \caption{Example of the Weighted Chain Arc Curve for each of six channels}
    \label{fig:multi_ac}
\end{figure}

Change point candidates are estimated based upon the intuition that actual change points should often coincide with either a local or global minimum in the $WCAC$. Figure~\ref{fig:multi_ac} shows $WCAC$, for six different time-series including a sequence of \textit{sitting, running and sitting} activities. The figure clearly illustrates that in each of the $WCAC$ channels, there are local minima in close proximity to the actual change points. These change point candidates are used as the search space for the entropy-based segmentation described in the next section.

\subsection{Entropy-based Segmentation.} \label{Entropy}
The shape-based segmentation is unable to detect segment boundaries of time-series that have non-repeating shape patterns. Furthermore, the representation can be biased due to the segment size variation across the time-series. Shorter segments are more likely to possess fewer arcs than longer segments, and hence, $WCAC$ can be biased towards estimating change points across the shorter segments. 
Consequently, we utilise the Information Gain (IG) metric to evaluate each local minima of $WCAC$ as a potential segment boundary. 
A greedy search procedure is implemented upon the set of boundary candidates in order to identify change points that minimize the entropy-based cost function in (\ref{eq:cost1}). Given the first term of the cost function ($H(X,\emptyset)$) is the entropy of the whole time-series as a single segment (no change-point), which have a constant value, 
maximising IG is equivalent to minimizing the entropy of the constituent segments. The cost function is defined as follows:


\begin{equation}
 \label{eq:cost1}
 \mathcal{L}= H(X,\emptyset)-\sum_{i}^{\left | \mathbf{S' \cup b} \right |+1} {\frac{\left | \mathbf{s}_{i} \right |}{\left | \mathbf{X} \right |}H(s_{i})}  
  \end{equation}
  
 \begin{equation}
 \label{eq:cost2}
 \maximize_{b \epsilon B} \mathcal{L}
  \end{equation}

where $X$ is the time-series of $d$ dimensions, $B$ is the set of segment boundaries that have been selected during previous shape-based segmentation, $s_i$ is the segment between the $i-1$ and $i$th selected boundaries, and $|.|$ is the length operator. $S'$ is the list of change points that are chosen by our greedy entropy based-method. $H(s_{i})$ is the Shannon entropy of the segment $i$:

\begin{equation}
 \label{eq:entropy}
    H(\mathbf{s}_{i}) = - \sum_{j=1}^{n}{p_{i}^{j} log p_{i}^{j}}
\end{equation}

and $p_{i}^{j}$ is the area of segment $s_{i}$ of series $X^{j}$ divided by this segment summed across all $d$ time-series. 

Algorithm \ref{alg:greedy} describes the greedy search used to estimate the segment boundaries (\textit{GreedyEntropySearch}), whilst Algorithm \ref{alg:greedyalgorithm} explains the complete \textit{ESPRESSO} approach. During each iteration of the $GreedyEntropySeg$, the remaining segment boundary candidates $b$ (not currently in $S'$) were used to split an existing segment of the time-series into two segments. The entropy of the new segments were then computed (line 6). The candidate $b$ that produced the two segments of lowest entropy were then selected and added to the set $S'$. This greedy search was repeated for each dimension of the time-series in $ESPRESSO$ (Algoirthm 4).



\begin{minipage}[t]{.49\textwidth}
\raggedleft
\begin{algorithm}[H]
\SetAlgoLined
\KwData{TS, candidtes}
\KwResult{Selected boundaries (S) and corresponding entropy (H)}

$TT, H, S$ = empty\;
\While{(!meetKneePoint(H))}{
$minH$ = Inf\;
\For{b in candidates}{
$TT$ = sort($TT$.append($b$))\;
$h$ = Entropy($TS$, $TT$)\;
\If { $h \leq minH$}{
$bestB = b$; $minH = h$\;
}
}
$S$.append($bestB$); $H$.append.($minH$)\;
$candidate$.remove($b$)\;
}
return S, H\;

\caption{GreedyEntropySeg}
 \label{alg:greedy}
\end{algorithm}
\end{minipage}
\begin{minipage}[t]{.49\textwidth}
\raggedright
\begin{algorithm}[H]
\SetAlgoLined
\KwData{TS}
\KwResult{TT, score}
    $MP, MPindex$ = calculateMatrixProfile($TS$)\;
    $WCAC$ = ExtractWCAC($MP, MPindex$)\;
    $candidates$ = findLocalMinima($WCAC$)\;
    
    \For {each dimenssion}{
    $TT[dim], entropy[dim]$ =  GreedyEntropySeg($TS$,$candidates$[$dim$])\;
    }
    $best\_index$ = IndexOfMin($entropy$)\; 
    return $TT$[$best\_index$], $entropy$[$best\_index$]\;
  
 \caption{ESPRESSO}
 \label{alg:greedyalgorithm}
\end{algorithm}
\end{minipage}
\subsection{Channel Ranking}

In this study, we showed that the segmentation accuracy was positively correlated with the entropy of the estimated segments in (\ref{eq:cost1}). Consequently, we attempt to exploit this discovery to rank the channels of our multi-dimensional time-series $X$ and select the channel that was most likely to provide a higher segmentation accuracy than $X$.  
For each channel $j$ in $X$, a set of boundary candidates ($B^{j}$) were estimated from its $WCAC^{j}$ representation. A greedy search of $B^{j}$ (as outlined in section \ref{Entropy}) was then used to segment $X$ based upon maximising the IG metric of (\ref{eq:cost1}). 

\subsection{Estimating the Number of Segments}
The number of segments were estimated by analyzing the extent to which a new segment contributes to decreasing the entropy of the segmented time-series. The relationship between entropy and the number of segments have been proven to be monotonic increasing \cite{sadri2017information}, given the entropy of constituent segments will always decrease as $k$ is increased. The following knee point detection equation in (\ref{eq:kneepoint}), which was proposed by \cite{sadri2017information}, was used to estimate $k$, where $\mathcal{L}_{k}$ denotes the reduced entropy for $k$ segments. 

\begin{equation} 
\label{eq:kneepoint}
\maximize_{k}(\frac{\mathcal{L}_{k} - \mathcal{L}_{k-1}}{\mathcal{L}_{k+1} - \mathcal{L}_{k}})\end{equation}



\section{Experiments And Evaluation}\label{sec:experiments}
We introduce the seven datasets and four benchmark segmentation methods used in our experiment. The metrics used to evaluate the performance of ESPRESSO and the benchmark segmentation methods are then defined. Finally, the results of the experiment and some use-case studies are presented. The source code is available at the GitHub page: \href{https://github.com/cruiseresearchgroup/ESPRESSO}{https://github.com/cruiseresearchgroup/ESPRESSO}.

\subsection{Datasets}
Seven public datasets comprised of smartphones, RFID tags and different wearable sensors including motion sensors, physiological sensors and eye wear computing sensors have been used to test the segmentation performance of \textit{ESPRESSO} and benchmark methods. Table 1 provides a detailed summary of the seven public datasets used in this experiment. 
We evaluated our method on two different types of wearable sensor datasets; continuous (C) and non-continuous (NC). The continuous datasets were comprised of sensor data collected across an uninterrupted sequence of different human activities. In these datasets, the activity transition times were manually recorded by human observers. The second type of datasets were comprised of individual recordings of human activity that were manually stitched together to form an activity sequence. Consequently, the transition between adjacent activities were far more discontinuous in the second type of datasets than the first type.
In addition, the datasets were categorised based on whether they contained repetitive temporal patterns (R) or were exclusively non-repetitive patterns (NR). Table \ref{tab:ds_category} presents the datasets and categories they are each associated with.\par

\begin{table}[]
\centering

\label{dataset_table}
\caption{Properties of the datasets}
\begin{tabular}{l|l|l|c|c|c}
\multicolumn{1}{c|}{Dataset}                              &
\multicolumn{1}{c|}{Context}&
\multicolumn{1}{c|}{Sensors}                                                                            & \multicolumn{1}{c|}{Length$^1$} &   Dimension$^2$ & Segments$^3$ \\ \hline
HandGesture\cite{bulling2014tutorial}& 10 hand gestures   & Accelerometer,Gyrometer   & 133.3K     & 18      & 600     \\

\hline
PAMAP\cite{reiss2012introducing,6083640}& 14 physical activities & Accelerometer,Temperature   & 42.4k      & 10      & 21      \\

\hline
USC-HAD \cite{zhang2012usc}  & 12 physical activities            & Accelerometer,Gyrometer& 93.6k      & 6       & 36      \\

\hline
EYE state\cite{rosler2013first}& close/open eye          & EEG  & 2k         & 8       & 5       \\

\hline
WESAD\cite{wesad2018} & \begin{tabular}[c]{@{}l@{}}Stress, amusement,\\ and meditation state\end{tabular} & \begin{tabular}[c]{@{}l@{}}ECG, EMG, EEG,\\Respiration\end{tabular}   & $\sim$300k & 4       & 8       \\

\hline
RFID\cite{yao2015unobtrusive}& 12 physical activities    & RFID Readers    & 15.3K      & 12      & 84      \\

\hline
Emotion\cite{HeinischHD18} & 
\begin{tabular}[c]{@{}l@{}}Neutral, negative\\and positive pleasure\end{tabular}  & 
\begin{tabular}[c]{@{}l@{}} EDA, EMG, BVP, ECG,\\ Temp,Respiration,Heartrate\end{tabular} & 
$\sim$37k  &   10      & 4   \\
\hline 
\end{tabular}
{\raggedright \par * $^1$ Length of each time-series, $^2$ Number of dimensions, and $^3$ Number of segments.\par}
\end{table}

\begin{table}[]
\centering
\caption{Categories of datasets.  
Datasets that contain both features are marked with \cmark;\xmark }
\begin{tabular}{c|ccccccc}
\backslashbox{Feature}{Dataset}           & PAMAP & RFID & Hand Gesture & USC-HAD & WESAD & EYE & Emotion \\ \hline
\begin{tabular}[c]{@{}c@{}} Continuous (\cmark) /\\ Non-Continuous (\xmark) segments\end{tabular}       & \xmark    & \xmark   & \cmark   &  \xmark & \xmark   &  \cmark  &   \cmark   \\
\begin{tabular}[c]{@{}c@{}} Repetitive (\cmark) /\\ Non-Repetitive (\xmark) patterns\end{tabular} &  \cmark;\xmark  &  \cmark;\xmark & \cmark;\xmark         & \cmark;\xmark     & \xmark    &  \xmark & \xmark
\end{tabular}
\label{tab:ds_category}

\end{table}

\paragraph{USC Human Activity Dataset (USC-HAD)}\footnote{http://sipi.usc.edu/had/}
\cite{zhang2012usc}: The USC-HAD dataset includes twelve human activities that were each recorded separately across fourteen subjects. Each human subject was fitted with a 3-axis accelerometer and a 3-axis gyrometer that was attached to the front of the right hip and sampled at 100Hz. Activities were repeated five times for each subject and consisted of: walking forward, walking left, walking right, walking upstairs, walking downstairs, running forward, jumping up, sitting, standing, sleeping, elevator up, and elevator down.
To perform experiments, the different set of activities were manually stitched randomly, therefore, USC-HAD was considered a NC dataset.\par

\paragraph{Physical Activity Monitoring for Aging People (PAMAP)}\footnote{http://www.pamap.org}
\cite{reiss2012introducing,6083640}: This dataset includes fourteen low-level (such as walking and sitting) and high-level (such as ironing which consists of two or more low-level) human activities undertaken by eight different subjects. Each subject was fitted with an IMU (inertial measurement units) sensor on their wrist, chest and ankle. Each participant was given both an indoor and outdoor activity schedule to perform the following activities sequentially: lying, sitting, standing, walking very slow, normal walking, Nordic walking, running, ascending stairs, descending stairs, cycling, ironing, vacuum cleaning, jumping rope and playing soccer.
Each IMU collected observations of temperature, 3-axis acceleration, 3-axis angular velocity (gyroscope), and the 3-axis magnetic field (magnetometer) at a sampling rate of 100Hz. As a result of missing readings in some of the sensors, only a subset of this IMU dataset was used in the experiment; the data from all three accelerometers and the hand fitted thermometer. 

\paragraph{Hand Gesture Dataset}\footnote{https://github.com/andreas-bulling/ActRecTut} \cite{bulling2014tutorial}: Our experiment used the Hand Gesture dataset, a collection of twelve hand movement activities performed by two subjects. Activities were captured by three IMUs that were attached to the subject's hand, upper arm and lower arm, respectively. The activities that were recorded within the experiments included: opening the window, closing the window, drinking, watering plants, cutting, chopping, stirring, reading a book, a tennis forehand, a tennis backhand and a tennis smash.
\par
\paragraph{Device-free posture recognition by RFID (RFID)} \cite{yao2015unobtrusive}:  In this experiment nine passive RFID tags were placed on a wall. The experimental dataset consist of six subjects that each performed twelve predefined postures between the wall and an RFID antenna. Each posture was performed for 60 seconds. RFID was a NC dataset given it was formed by concatenating the twelve postures for each of the six subjects.\par

\paragraph{Emotion Dataset}\footnote{https://www.comtec.eecs.uni-kassel.de/emotiondata/} \cite{HeinischHD18,heinisch2019angry}: This dataset has been collated to study the physiological response to different emotional states and to identify the effect of physical activity on these emotion state. Five hours of physiological data from E4-wristband, Biosignalsplux device, and smart-phone were collected from 18 subjects with respect to three different emotion categories, High Positive Pleasure High Arousal (HPHA), High Negative Pleasure High Arousal (HNHA), and Neutral. 

\paragraph{WESAD Dataset}\footnote{https://ubicomp.eti.uni-siegen.de/home/datasets/icmi18} \cite{wesad2018}: WESAD is a well-known dataset for stress that has been acquired from multi-modal wearable sensors. This dataset is comprised of physiological and motion data from chest and wrist-worn sensors of 15 subjects. In this study, only chest-worn sensors with a down-sampling rate of 10 were used to detect stress, amusement and meditation segments.

\paragraph{EYE State Dataset}\footnote{https://archive.ics.uci.edu/ml/datasets/EEG+Eye+State} \cite{rosler2013first}: This dataset consists of 14980 samples of 15 EEG sensors collecting eye state data for 117 seconds. The labels (close/open) are manually annotated using video collected during the measurements.\par
 
\subsection{Benchmark methods}
 The performance of the proposed \textit{ESPRESSO} method was compared to four state-of-the-art algorithms: Fast Low-Cost Semantic Segmentation (\textit{FLOSS}) \cite{gharghabi2018domain}, Information Gain-based Time-series Segmentation (\textit{IGTS}) \cite{sadri2017information}, additive Hilbert-Schmidt Independence Criterion (\textit{aHSIC}) \cite{Yamadaahsic}, and Relative unconstrained Least Square Importance Fitting (\textit{RuLSIF}) \cite{liu2013change}. 
To avoid inconsistencies and implementation errors, we evaluated our method against benchmark algorithms with publicly available source code.\par

\textit{FLOSS} is a shape-based segmentation method that sits on top of the Matrix Profile time-series representation, and \textit{IGTS}, is an Information gain based segmentation method, which have both been described in previous sections. \textit{IGTS} requires no input parameters, whilst \textit{FLOSS} requires the subsequence length as its input parameter. \textit{RuLSIF} is based on estimating the relative probability density ratio of subsequences. The number of subsequences in each round and regularization constant were fixed at 10 and 0.01, respectively, as suggested by the authors. To enable a fair comparison with this method, we evaluate its performance across different subsequence lengths.
The final method, \textit{aHSIC}, is a multi-dimensional time-series segmentation combined with channel selection. Firstly, this method selects important channels using a supervised learning method and then scores each time step according to the proposed dependency measure regarding a pseudo-binary-label. The regularization constant and the kernel parameter was set as 0.01 and 1, respectively, as suggested in their paper. 
Segmentation performance was compared across a range of subsequence sizes that were unique to each dataset based on its sample rate and the minimum segment duration of 0.5 seconds. The range of subsequence sizes varied from the narrowest set of 10 to 40 samples for the RFID dataset to the widest set of 100 to 900 samples for the PAMAP dataset.

\subsection{Evaluation Metrics}\label{eval}
The performance of the segmentation algorithms were
evaluated with respect to the following metrics:
 
\begin{enumerate}
\item F-score: the F-score is defined as the harmonic mean of the Precision ($\frac{TP}{TP+FP}$) and Recall ($\frac{TP}{TP+FN}$). Each estimated segment was defined as a True Positive ($TP$) when it was located within a specified time window of the ground truth segment boundaries and a False Negative ($FN$) when it fell outside the time window of all the ground truth segment  boundaries. When multiple segment estimates fell within a specified time window of the ground truth segment boundary, only the closest estimate was considered to be $TP$ and the remaining estimates were considered to be False Positives ($FP$). As a consequence of the sampling rate of sensors in the dataset, the time window (i.e. segmentation threshold) was set to 0.5 seconds for the EYE dataset and 2 seconds for each of the remaining datasets.


    \item RMSE: The Root Mean Square Error (RMSE) was computed between the ground truth segment boundary time and its nearest estimated segment boundary time. The RMSE was then normalized into the range of [0, 1] by dividing it by the time-series duration.\par
    
    \item MAE: To compare the performance of the proposed shape-based segmentation method, $WCAC$, and the other shape-based segmentation method, \textit{FLOSS}, we employed the Mean Absolute Error (MAE) as used in \cite{gharghabi2017matrix}. For this particular study, segmentation performance was evaluated as the MAE between the estimated segment boundaries and ground truth segment boundaries. 
\end{enumerate}

The F-score metric depends upon the selection of a threshold value. For example, consider the actual segment transition time is at 250 seconds and a threshold of five seconds is set. Suppose two segmentation methods estimate segments boundaries, A and B, at 256 seconds and 300 seconds, respectively. The F-score metric will evaluate the A and B estimates as False Negatives despite A being a far superior estimate to B. The RMSE and MAE, will address this problem given their continuous metric space ensures A will be represented as a superior estimate to B.  
Typically all metrics might incorporate the error of several transition estimates that are in closest proximity to a single ground truth boundary. Consequently, we ensure each ground truth boundary is exclusively mapped to only one segment boundary estimate to ensure that metrics are not biased by change point estimates being clustered around a subset of segment boundaries.
Existing studies often evaluate performance using one of these metrics, however, we believe including the F-score as well as RMSE (or MAE) provides a more comprehensive evaluation.






\subsection{Evaluation}
In this section, we first compare the effectiveness of the proposed shape-based segmentation technique, $WCAC$ against \textit{FLOSS}. Then, we investigate the performance of \textit{ESPRESSO} against four state-of-the-art segmentation techniques. Finally, we undertake an ablation study to compare the performance of the shape and entropy based components of \textit{ESPRESSO}.

\subsubsection{$WCAC$} 
\begin{figure}[]
\centering
\begin{minipage}[t]{0.48\linewidth}
\subfigure{
\label{fig:ac_compare_hg}
  \includegraphics[width=\linewidth, height=1.8in]{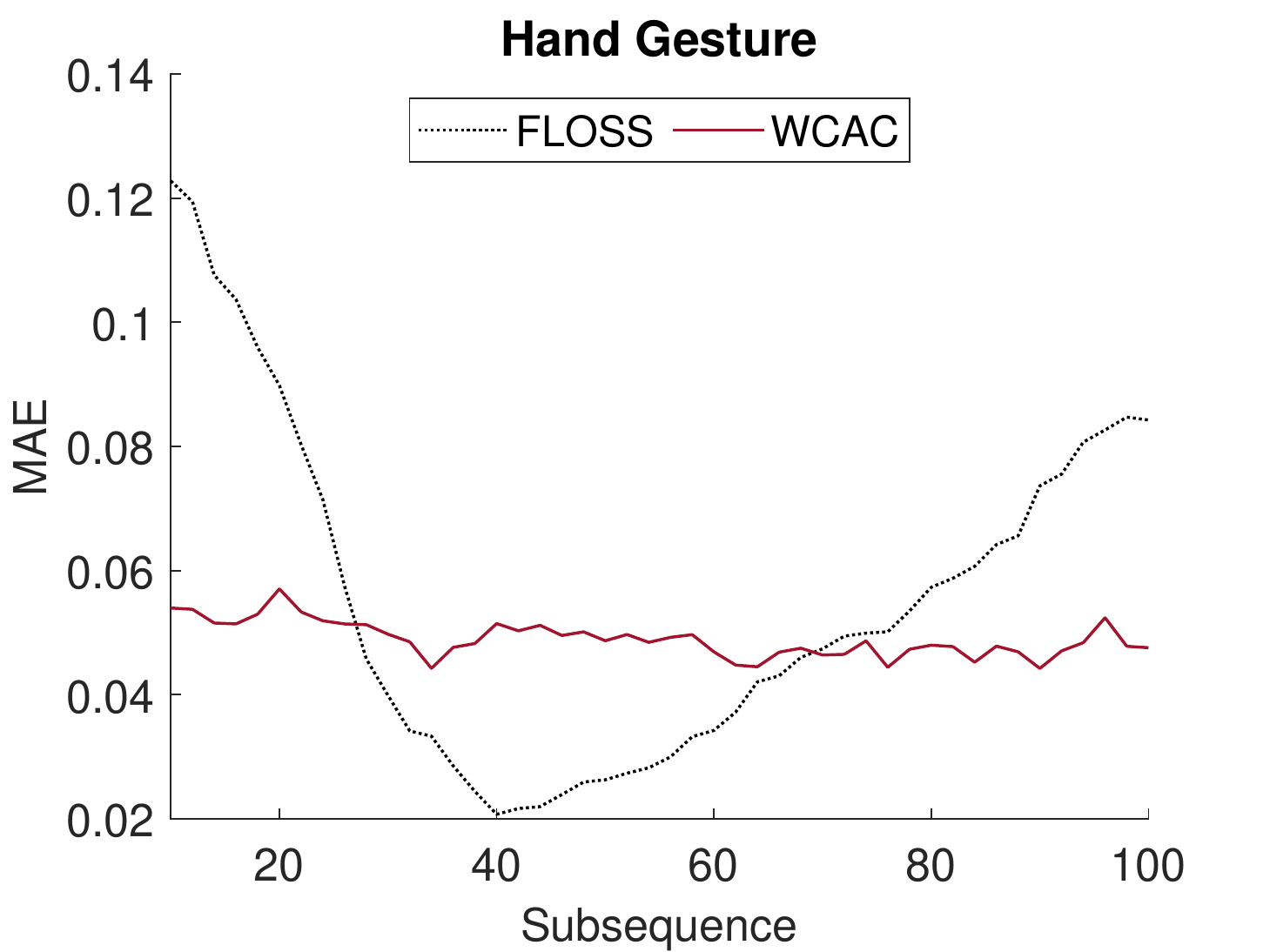}}

\subfigure{
\label{fig:ac_compare_usc}
  \includegraphics[width=\linewidth, height=1.8in]{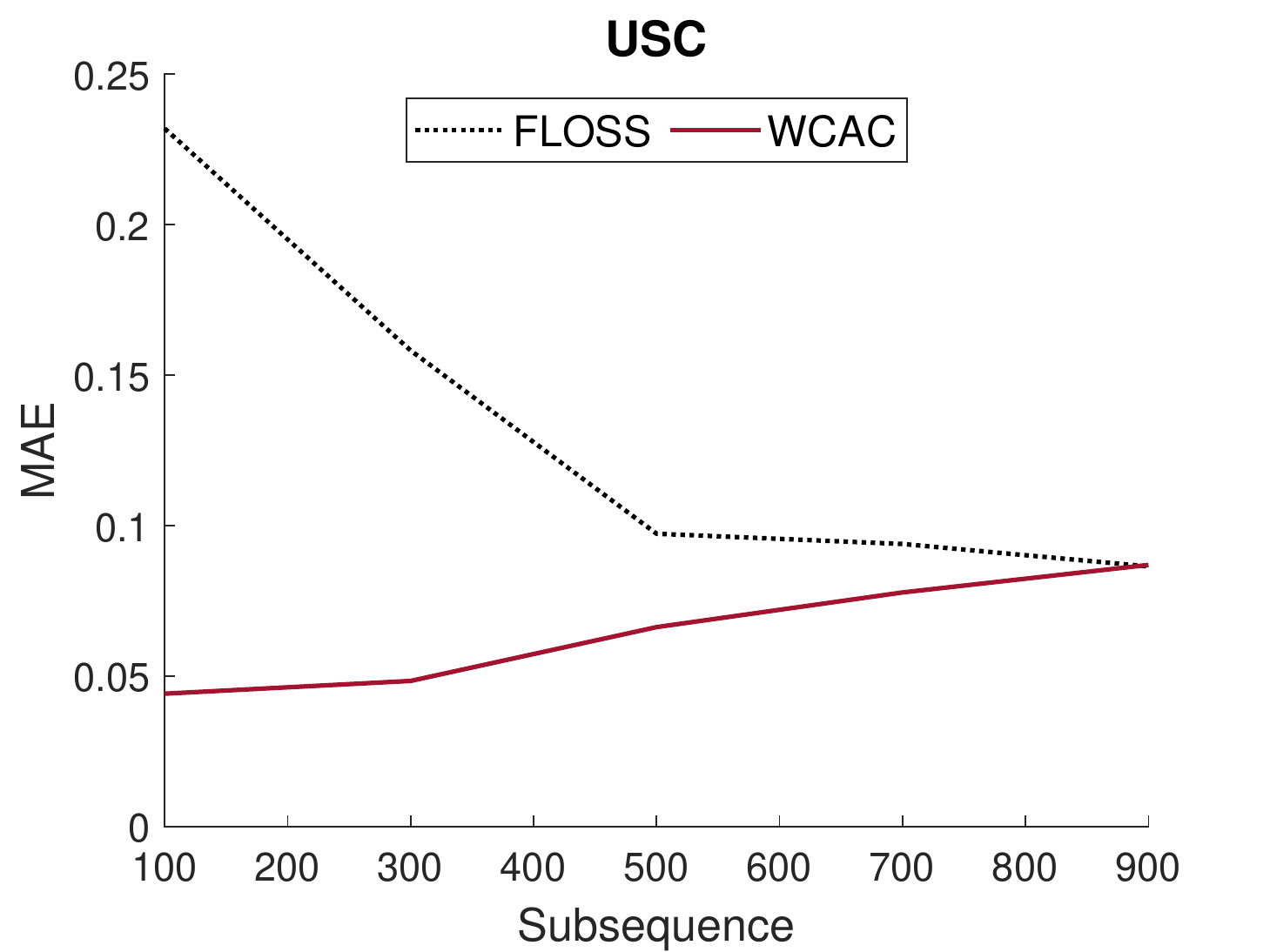}}

\subfigure{
\label{fig:ac_compare_rfid}
  \includegraphics[width=\linewidth, height=1.8in]{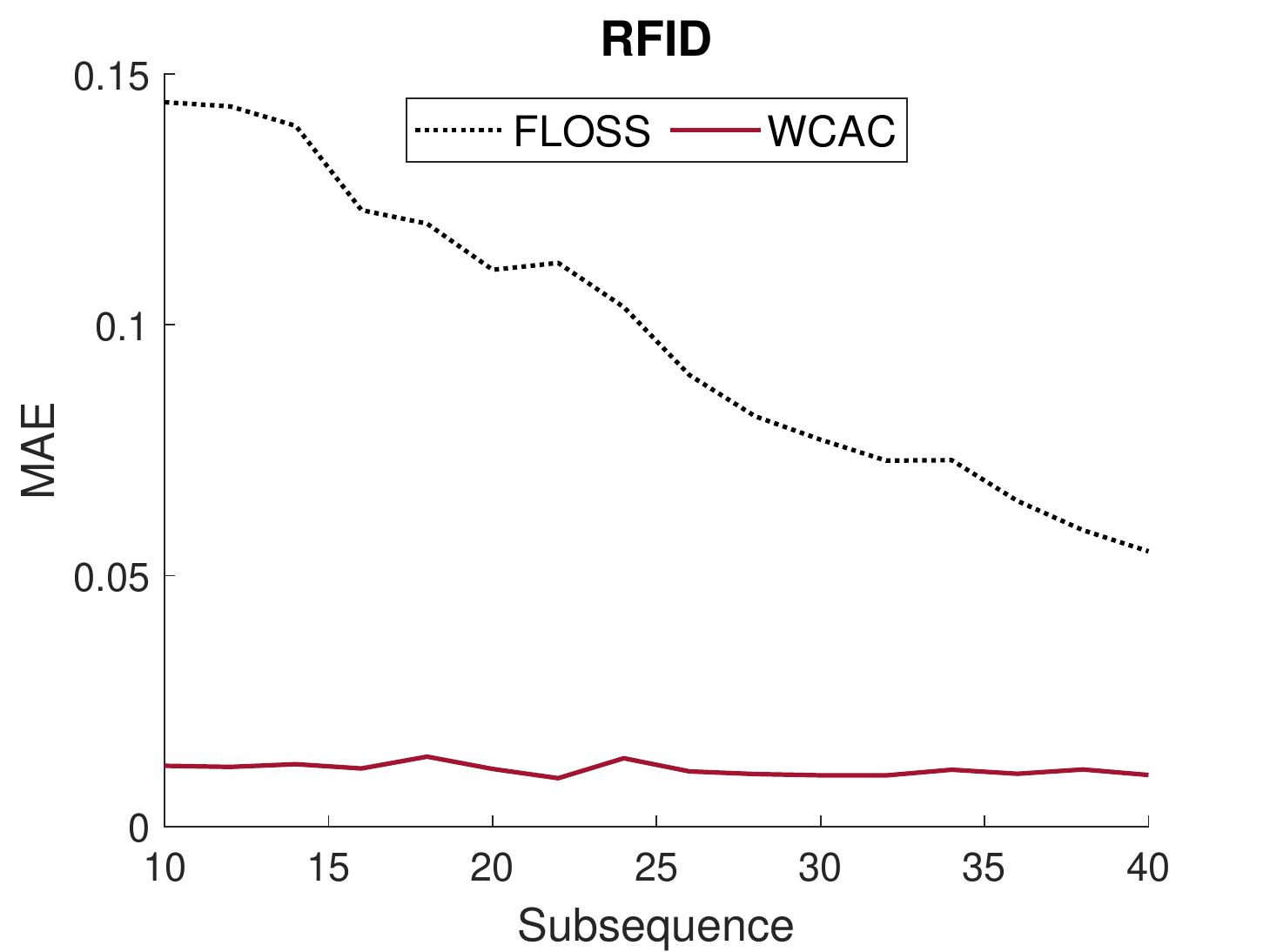}}
\caption[]{Comparing the performance of \textit{FLOSS} and the proposed weighted chained arc curve ($WCAC$) for the \subref{fig:ac_compare_hg} Hand gesture, 
\subref{fig:ac_compare_usc} USC-HAD, and
\subref{fig:ac_compare_rfid} RFID datasets.}
\label{fig:ac_compare}
\label{fig:minipage1}
\end{minipage}
\quad
\begin{minipage}[t]{0.48\linewidth}
\subfigure{
\label{fig:hg_rmse}
  \includegraphics[width=\linewidth, height=1.8in]{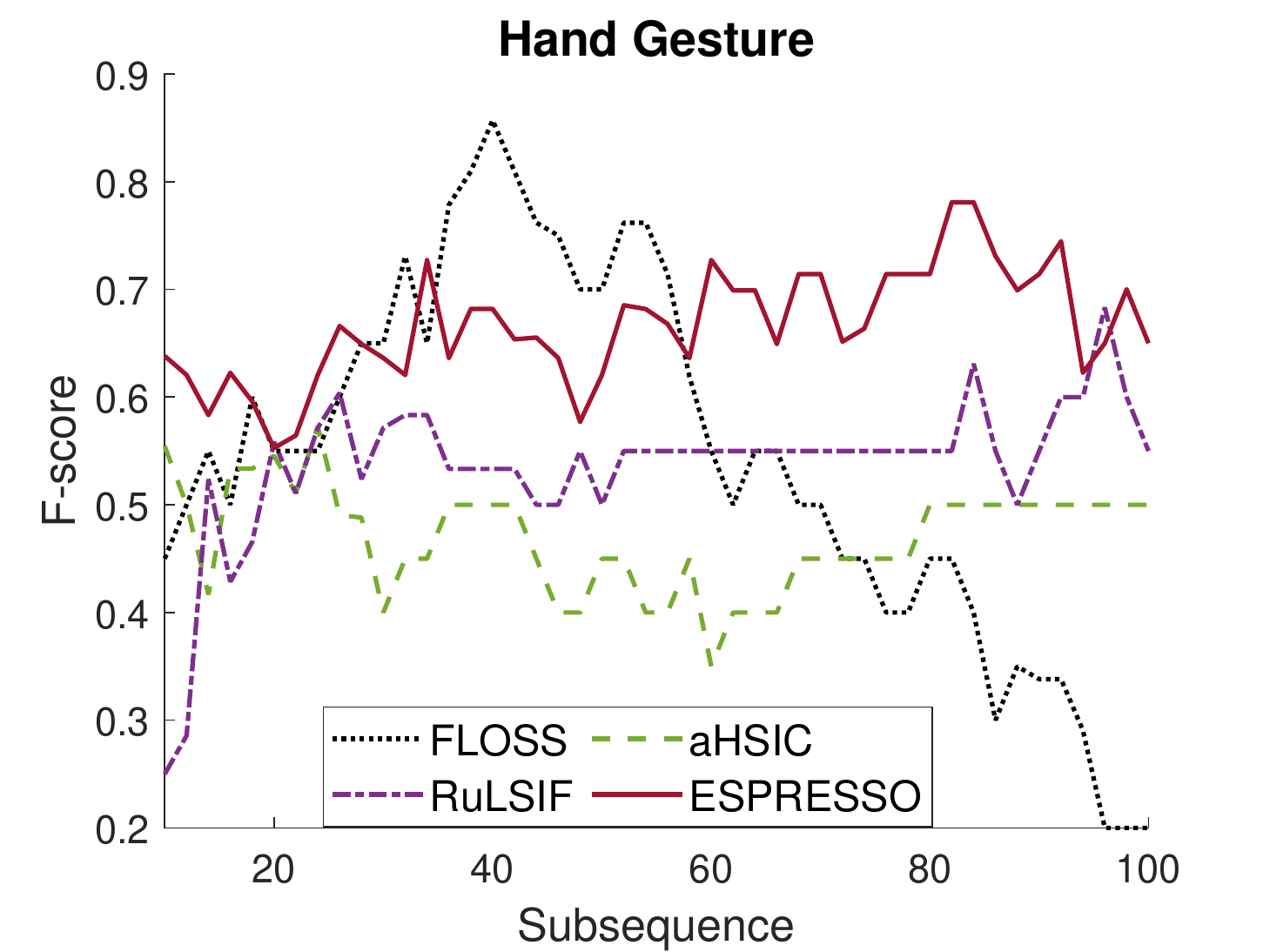}}

\subfigure{
\label{fig:pamap_rmse}
  \includegraphics[width=\linewidth, height=1.8in]{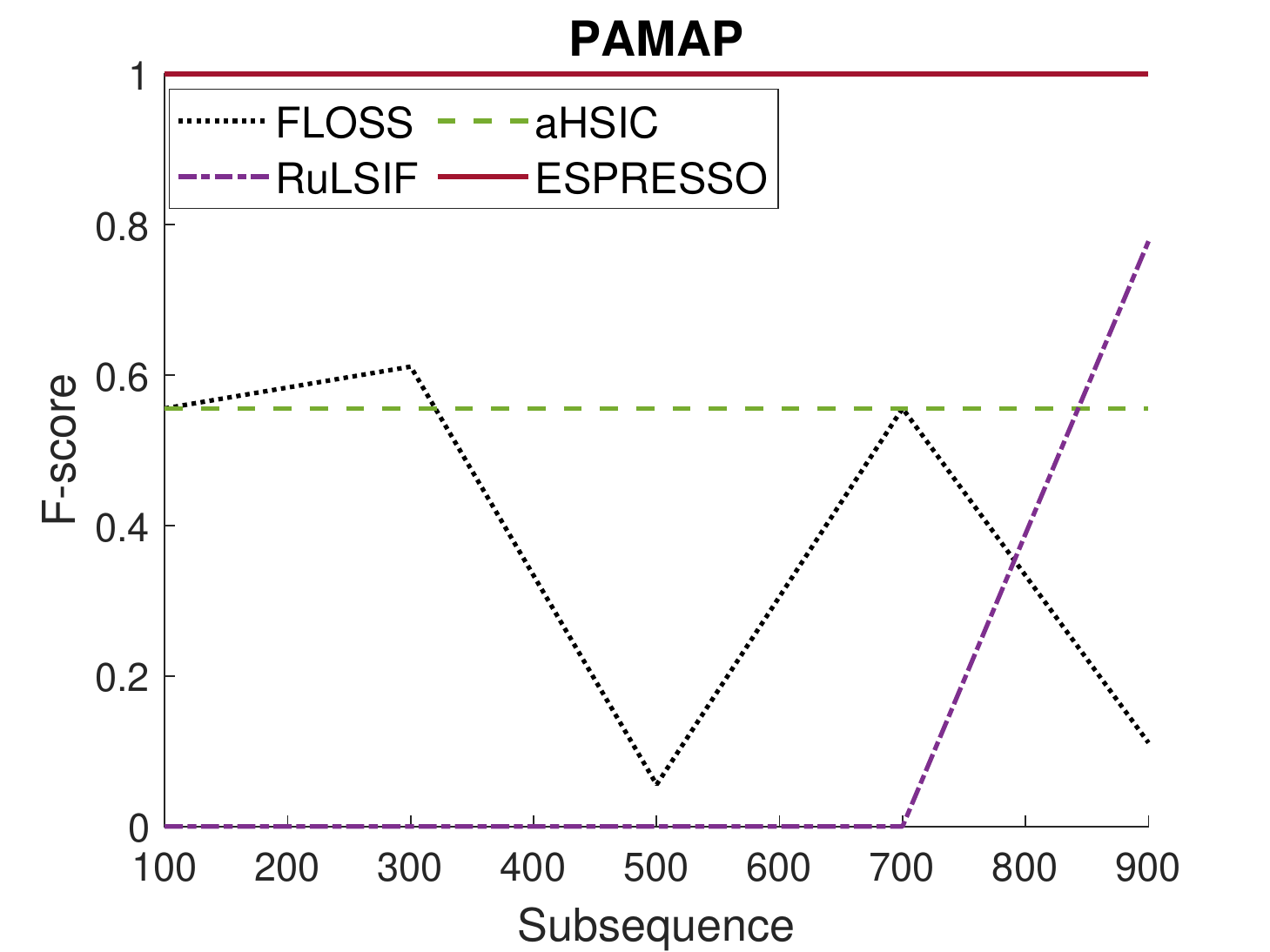}}

\subfigure{
\label{fig:rfid_rmse} \includegraphics[width=\linewidth, height=1.8in]{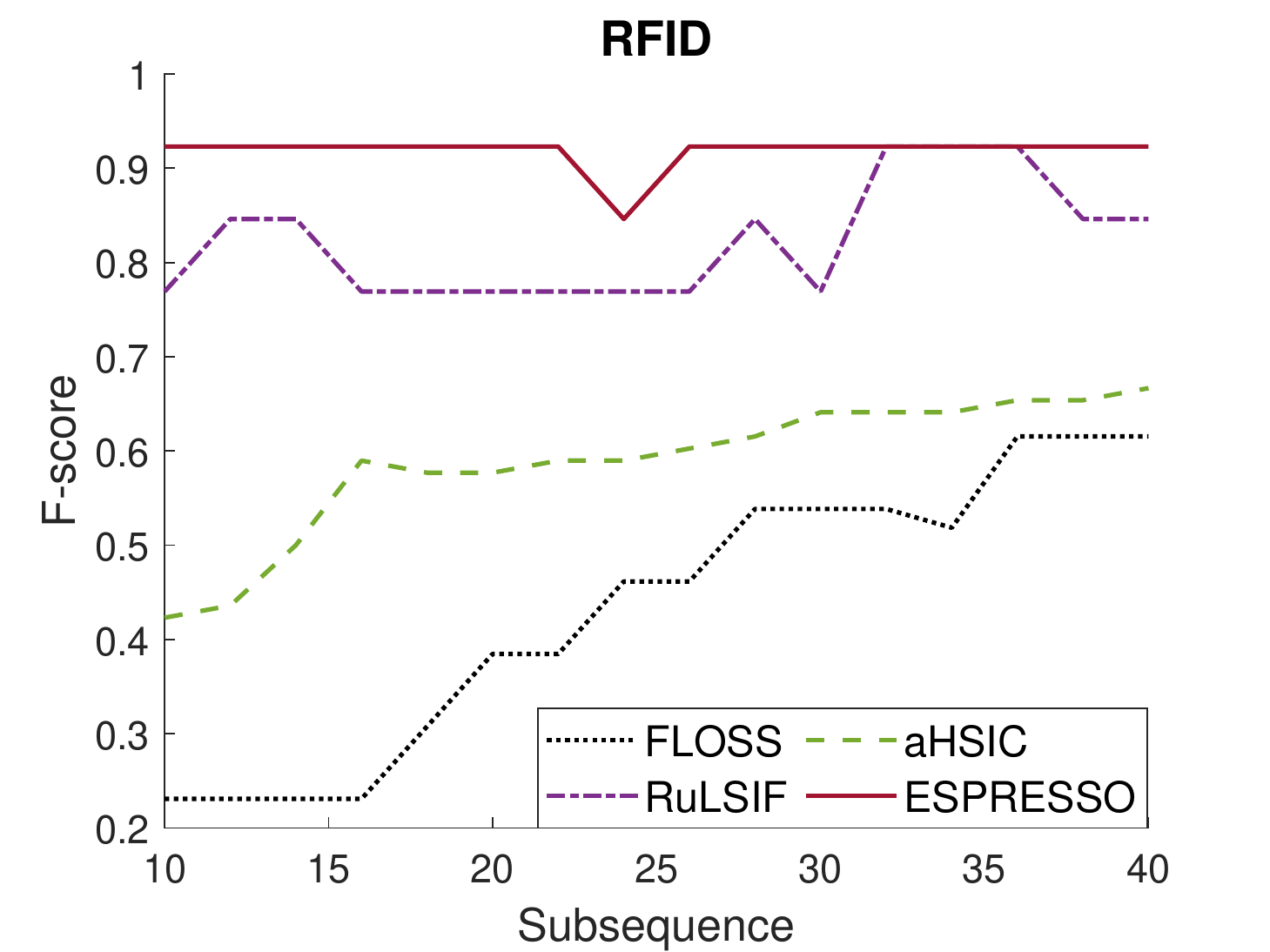}}
\caption{The segmentation performance (F-score) of the \textit{ESPRESSO} method and three benchmark methods across a range of subsequence lengths for three datasets (a) Hand Gesture (b) PAMAP and (c) RFID}
  \label{fig:plots}

\end{minipage}
\end{figure}

In this section, we compare the performance of our proposed shape-based segmentation method, $WCAC$, and state of the art shape based segmentation method \textit{FLOSS}. To compare the effectiveness of these methods, the Hand Gesture, USC-HAD and RFID datasets were selected given they contained a diverse set of sensors and contained repetitive temporal patterns. The experiments were repeated over a set of subsequence lengths ranging from between 10 and 40 samples for the Hand Gesture dataset \ref{fig:ac_compare_hg}, 50 and 550 samples for the USC-HAD dataset \ref{fig:ac_compare_usc} and 20 and 100 samples for the RFID dataset \ref{fig:ac_compare_rfid}. The subsequence lengths were set based upon the sampling rate of each dataset. The minimum subsequence length was set to 0.5 seconds, whilst the maximum subsequence length was set to half of the minimum segment size in the dataset. 

Figure \ref{fig:ac_compare} shows the segmentation performance with respect to the Mean Absolute Error (MAE). These figures show that the proposed $WCAC$ method does not only have consistently superior segmentation performance to \textit{FLOSS}, but was far more insensitive to the subsequence length that was used.\par

\noindent

\subsubsection{ESPRESSO}\label{chap:ESPRESSO}
The performance of \textit{ESPRESSO} was compared to four competing segmentation techniques: \textit{FLOSS}, \textit{IGTS}, \textit{aHSIC}, and \textit{RuLSIF}. We performed extensive experiments across seven datasets and over a range of different subsequence lengths. 
Figure ~\ref{fig:plots} compares the F-score of the \textit{ESPRESSO}, \textit{aHSIC}, \textit{RulSIF} and \textit{FLOSS} methods with respect to the subsequence length across three datasets; Hand Gesture, PAMAP and RFID. The \textit{IGTS} method was not included in this comparison given it does not utilise subsequences in order to perform segmentation. 

For the PAMAP and RFID datasets,
\textit{ESPRESSO} was shown to offer a very high level of segmentation performance that was superior to three benchmark algorithms across all subsequence lengths, apart from a small set of subsequence lengths (with 33 to 37 samples) in the RFID dataset where aHSIC's performance was equivalent. In addition, ESPRESSO's performance was consistently high across all subsequence lengths, a desirable property of the algorithm. 

For the Hand Gesture dataset, \textit{ESPRESSO} achieved superior segmentation performance to
the benchmark methods for a large majority of the subsequence lengths. The exception was a narrow range of subsequence lengths (with 35 to 55 samples) where \textit{FLOSS} outperformed \textit{ESPRESSO}. \textit{FLOSS}'s performance was found to be far more sensitive to subsequence length than \textit{ESPRESSO}, given it exhibited a significant performance decline outside of this optimal subsequence range.



Table \ref{tab:result} shows the average F-score and RMSE of \textit{ESPRESSO} and four benchmark methods across each of the seven datasets. For each of the methods apart from \textit{IGTS}, the F-score and RMSE were averaged across all of the subsequence lengths and subjects of each dataset. For \textit{IGTS}, the F-score was only averaged across the dataset subjects given its a top-down method that does not utilise subsequences. In the PAMAP and WESAD datasets, given the computational cost of \textit{aHSIC} became prohibitively high over longer window sizes, the window size was fixed at 50 samples as suggested in their paper. 

The segmentation results in Table \ref{tab:result} indicate that \textit{ESPRESSO} was superior to the benchmark methods, on average, across all datasets with an F-Score improvement of 45.6\%, 7\%, 44.4\%, and 45.2\% over the \textit{FLOSS, IGTS, aHSIC}, and \textit{RuLSIF} methods, respectively. In addition, \textit{ESPRESSO} had an average RMSE improvement of 140\%, 21\%, 92\% and 224\% over the \textit{FLOSS, IGTS, aHSIC}, and \textit{RuLSIF} methods, respectively. 
\textit{ESPRESSO} had a clear advantage over the \textit{FLOSS, RulSIF, aHSIC} methods across each of the datasets.
The RFID dataset was the only one where \textit{IGTS} was shown to be superior to \textit{ESPRESSO} across both performance metrics. \textit{ESPRESSO} was advantageous over \textit{IGTS} across four of the seven datasets (RFID, Hand Gesture, USC-HAD and WESAD) given both of its performance metrics were superior. We hypothesize that RFID was an optimal dataset for a top-down entropy based approach such as \textit{IGTS}, given the segments had salient statistical differences (as shown in Figure \ref{fig:example1}(b)).


\begin{table}[t]
\centering
\caption{Evaluation results}
\label{tab:result}
\begin{tabular}{c|l|ccccccc}
\multicolumn{1}{l}{}         & \backslashbox{Feature}{Dataset}    & PAMAP      & RFID          & Hand Gesture  & USC-HAD       & WESAD         & EYE           & Emotion        \\ 
\hline
\multirow{5}{*}{\rotatebox[origin=c]{90}{F-score}} 
& IGTS     & 1              & \textbf{0.9554}   & 0.3825          & 0.7333          & 0.6154          & 0.5116          & 0.5556           \\
& FLOSS    & 0.3778         & 0.4106            & 0.5379          & 0.3733          & 0.1795          & 0.4252          & 0.4722           \\
& aHSIC    & 0.5556         & 0.7787            & 0.2188          & 0.4             & -               & 0.5312          & 0.00                \\
& RuLSIF   & 0.1556         & 0.8560            & 0.2529          & 0.4133          & 0.3667          & 0.5336          & 0.2222            \\
& ESPRESSO & \textbf{1}     & 0.9378            & \textbf{0.6209} & \textbf{0.7467} & \textbf{0.6410} & \textbf{0.5821} & \textbf{0.5833}  \\ 

\hline
\multirow{5}{*}{\rotatebox[origin=c]{90}{RMSE)}}    
& IGTS     & \textbf{0.0024} & \textbf{0.0401} & 0.4270          & 0.1939          & 0.2195          & 0.0997          & \textbf{0.0607}           \\
& FLOSS    & 0.2779          & 0.3969          & 0.3166          & 0.3267          & 0.5140          & 0.1114          & 0.1219           \\
& aHSIC    & 0.1659          & 0.1411          & 0.4069          & 0.3147          & -               & 0.1070          & 0.2359            \\
& RuLSIF   & 0.8375          & 0.1013          & 0.3792          & 0.3338          & 0.2873          & 0.2727          & 0.5746            \\
& ESPRESSO & 0.0030          & 0.0692          & \textbf{0.2764} & \textbf{0.1933} & \textbf{0.1936} & \textbf{0.05}   & 0.0719

\end{tabular}
\end{table}

 \noindent
\begin{figure}[]
\centering
\begin{minipage}[t]{0.48\linewidth}
\subfigure{
\label{fig:ds-cat-fscore}
  \includegraphics[width=\linewidth, height=1.8in]{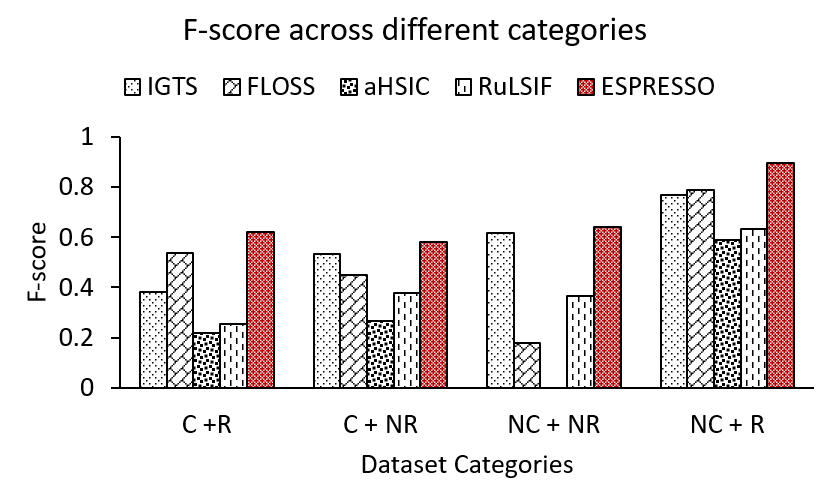}}

\end{minipage}
\quad
\begin{minipage}[t]{0.48\linewidth}
\subfigure{
\label{fig:ds-cat-rmse}
  \includegraphics[width=\linewidth, height=1.8in]{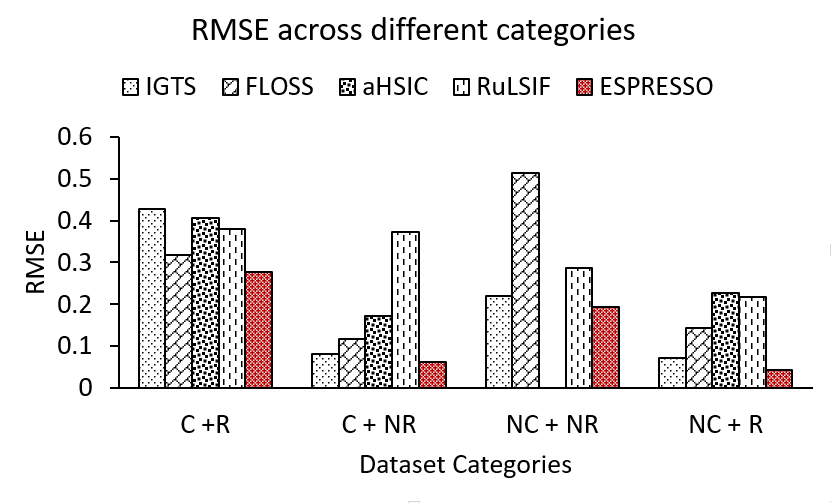}}

\end{minipage}
\caption{A comparison of \textit{ESPRESSO} and four benchmark methods across  different dataset categories; continuous datasets with repetitive patterns (C+R), continuous datasets with non-repetitive patterns (C+NR), non-continuous datasets with non-repetitive patterns (NC+NR) and non-continuous datasets with repetitive patterns (NC+R) (F-score(left) and RMSE(right)).}
  \label{fig:ds-categories-figs}
\end{figure}


The effectiveness of \textit{ESPRESSO} was then examined with respect to particular dataset characteristics in Figure \ref{fig:ds-categories-figs}:
\begin{itemize}
    \item \textbf{Datasets with continuous (C) and non-continuous (NC) segments:} 
    the datasets associated 
    with the C and NC categories are shown in Table \ref{tab:ds_category}.
    \textit{ESPRESSO} was shown to be superior to each of the benchmark methods for the C and NC categories with an average F-score of 0.59 and 0.67, respectively. Figure \ref{fig:ds-categories-figs} shows \textit{$ESPRESSO$'s} performance advantage over the statistical based methods ( \textit{IGTS}, \textit{RulSIF}) was more significant for the continuously recorded datasets than the non-continuous datasets. This suggests \textit{ESPRESSO} was better equipped to detect segment transitions with higher correlation than statistical methods. In contrast, \textit{ESPRESSO's} performance advantage over the shape-based \textit{FLOSS} method was greater for the NC datasets, in particular, the NC datasets with non-repeating temporal patterns.

    

    \item \textbf{Datasets with repeating temporal patterns (R) and non-repeating temporal patterns (NR):}
    HAR datasets consist of physical activities that contain both repeating (such as walking or stirring) and non-repeating (such as sitting or opening the window) actions. The association between the seven datasets used in the experiment and action categories are shown in Table \ref{tab:ds_category}.
    In Figure \ref{fig:ds-categories-figs}, \textit{ESPRESSO}'s performance was shown to be far superior across datasets that possess at least some repeated patterns (R) (average F score of 0.83) when compared to the datasets composed of non-repeating patterns (NR) exclusively (average F score of 0.6). This can be attributed to \textit{ESPRESSO} using $WCAC$ to identify potential segment boundary candidates; $WCAC$ is far more effective at detecting segment transitions across time-series with repeating shape patterns. 
    \textit{ESPRESSO} was shown to be superior to each of the four benchmark methods across both categories of temporal patterns. Figure \ref{fig:ds-categories-figs} shows \textit{ESPRESSO's} performance advantage was greater across R datasets than NR datasets. This could be attributed to ESPRESSO's unique ability to exploit shape and statistical properties of time-series in the R datasets where both properties are useful to segment the combination of repetitive and non repetitive patterns.

    \par

\end{itemize}


\subsubsection{Ablation Study}
\begin{figure}[b]
\centering
\begin{minipage}[t]{0.48\linewidth}
\subfigure{
\label{fig:ablation-fscore}
  \includegraphics[width=\linewidth, height=1.8in]{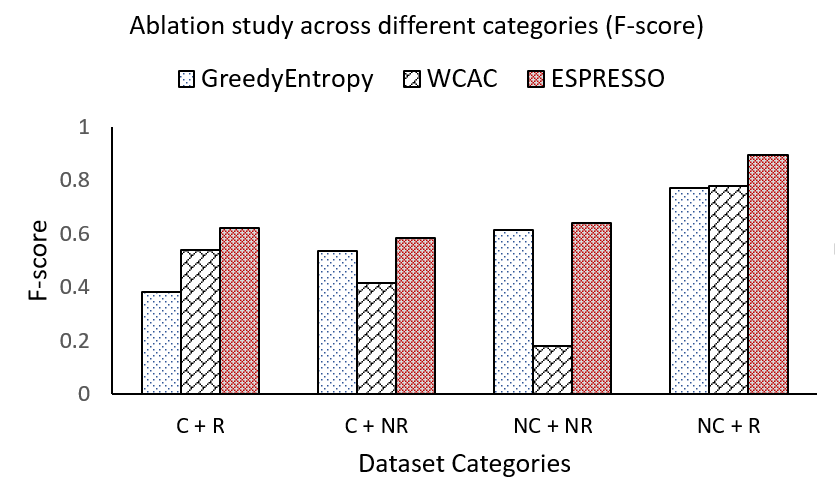}}

\end{minipage}
\quad
\begin{minipage}[t]{0.48\linewidth}
\subfigure{
\label{fig:ablation-rmse}
  \includegraphics[width=\linewidth, height=1.8in]{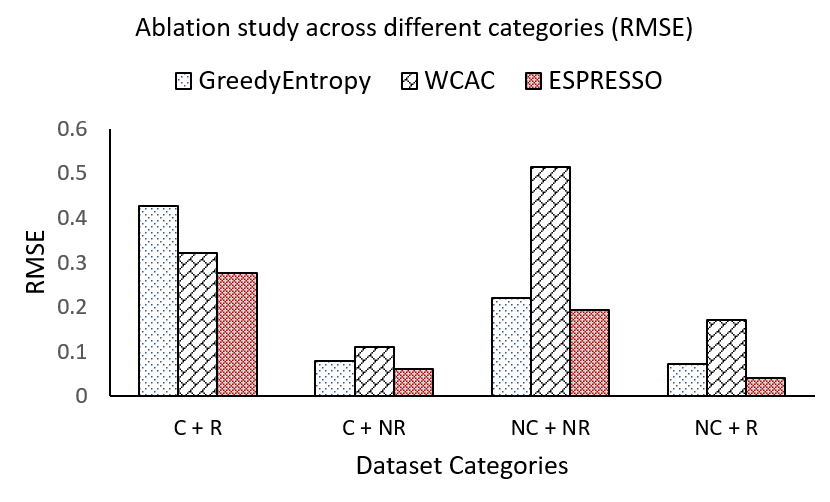}}

\end{minipage}
\caption{Investigating the contribution of shape and entropy based methods of $ESPRESSO$ across: continuous datasets with repetitive patterns (C+R), continuous datasets with non-repetitive patterns (C+NR), non-continuous datasets with non-repetitive patterns (NC+NR) and non-continuous datasets with repetitive patterns (NC+R) (F-score(left) and RMSE(right)). }
  \label{fig:ablation-figs}
\end{figure}

 An additional study is performed to investigate the temporal shape and statistical components of \textit{ESPRESSO} independently. Table \ref{tab:ablation} compares the segmentation performance of \textit{ESPRESSO} with its constituent shape-based method ($WCAC$) and entropy-based method (\textit{GreedyEntropySeg}). Furthermore, Figure \ref{fig:ablation-figs} compares the segmentation performance of the $WCAC$ and \textit{GreedyEntropySeg} methods across the four dataset categories introduced in section \ref{chap:ESPRESSO}. 
  
  The segmentation performance of \textit{ESPRESSO} was largely attributed to the strong and consistent segmentation performance of the \textit{GreedyEntropySeg} method with respect to the NC and NR dataset categories. In contrast, the $WCAC$ method was shown to offer a more significant contribution to \textit{ESPRESSO}'s segmentation performance for the R dataset categories (with repeating patterns). This can be attributed to $WCAC$ needing to exploit repeated temporal shape patterns in order to perform accurate segmentation of the time-series.

\begin{table}[h]
\centering
\caption{Ablation study}
\begin{tabular}{c|l|ccccccc}
\multicolumn{1}{l}{}         & \backslashbox{Feature}{Dataset}    & PAMAP      & RFID          & Hand Gesture  & USC-HAD    & WESAD         & EYE           & Emotion        \\ 
\hline
\multirow{3}{*}{\rotatebox[origin=c]{90}{F-score}} 
& WCAC              & 0.4218    & 0.5237           & 0.5381         & 0.3881         & 0.1795         & 0.4404          & 0.3889 \\
& GreedyEntropySeg. & 1        & \textbf{0.9554}  & 0.3825         & 0.7333         & 0.6154         & 0.5116          & 0.5556 \\
& $ESPRESSO^{*}$      &\textbf{1} & 0.9378           & \textbf{0.6609}& \textbf{0.7467}& \textbf{0.6410}& \textbf{0.5325} & \textbf{0.5833}  \\  
\hline
\multirow{3}{*}{\rotatebox[origin=c]{90}{RMSE}}   
& WCAC              & 0.1977          & 0.2528          & 0.3221         & 0.4109         & 0.5140         & 0.1         & 0.1222 \\
& GreedyEntropySeg. & \textbf{0.0024} & \textbf{0.0401} & 0.4270         & 0.1939         & 0.2195         & 0.0997          & \textbf{0.0607}   \\
& $ESPRESSO^{*}$      & 0.0030          & 0.0692          & \textbf{0.2764}& \textbf{0.1933}& \textbf{0.1936}& \textbf{0.05} & 0.0719    \\   
\hline
\end{tabular}
{\raggedright \par * Combination of WCAC + GreedyEntropy Segmentation.\par}
\label{tab:ablation}

\end{table}


\section{Real World Case Studies}
\subsection{Case Study1: Unsupervised Inference of Work Routines and Behavior Deviations }
In this section we show how performing segmentation with the data of wearable sensors can help to extract a user’s daily life patterns and to identify any deviations in their daily routines. In this study, we used an existing life-logging dataset from NTCIR-13 Life-logging track \cite{gurrin2019test}. 
 The data from three biometric sensors (calories burnt, heart rate, and skin temperature) and a step counter were used in this study. 
The aim of this study was to model the physical activity intensity of a user on a daily basis and then compare these in order to detect any activity deviations \cite{deldari2016inferring}. 
To extract different levels of activity intensity, we changed the granularity of the estimated segments from 4 segments to 11 segments per day.
 \noindent
\begin{figure}[b]
\centering
  
  \subfigure[][]{%
\label{fig:wsdm4}%
  \includegraphics[width=0.49\linewidth]{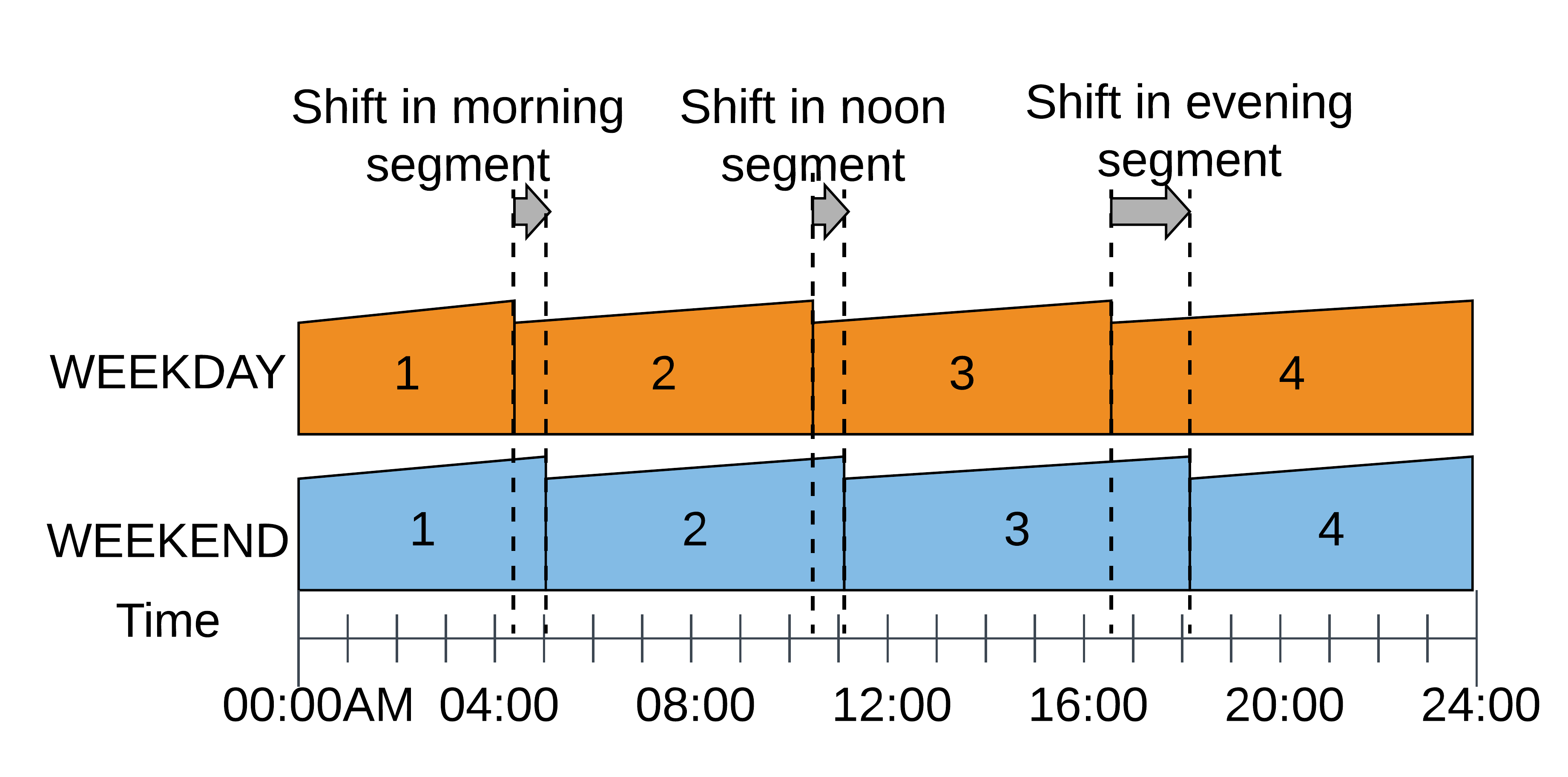}}%
  \hspace{1pt}%
\subfigure[][]{%
\label{fig:wsdm7}%
  \includegraphics[width=0.49\linewidth]{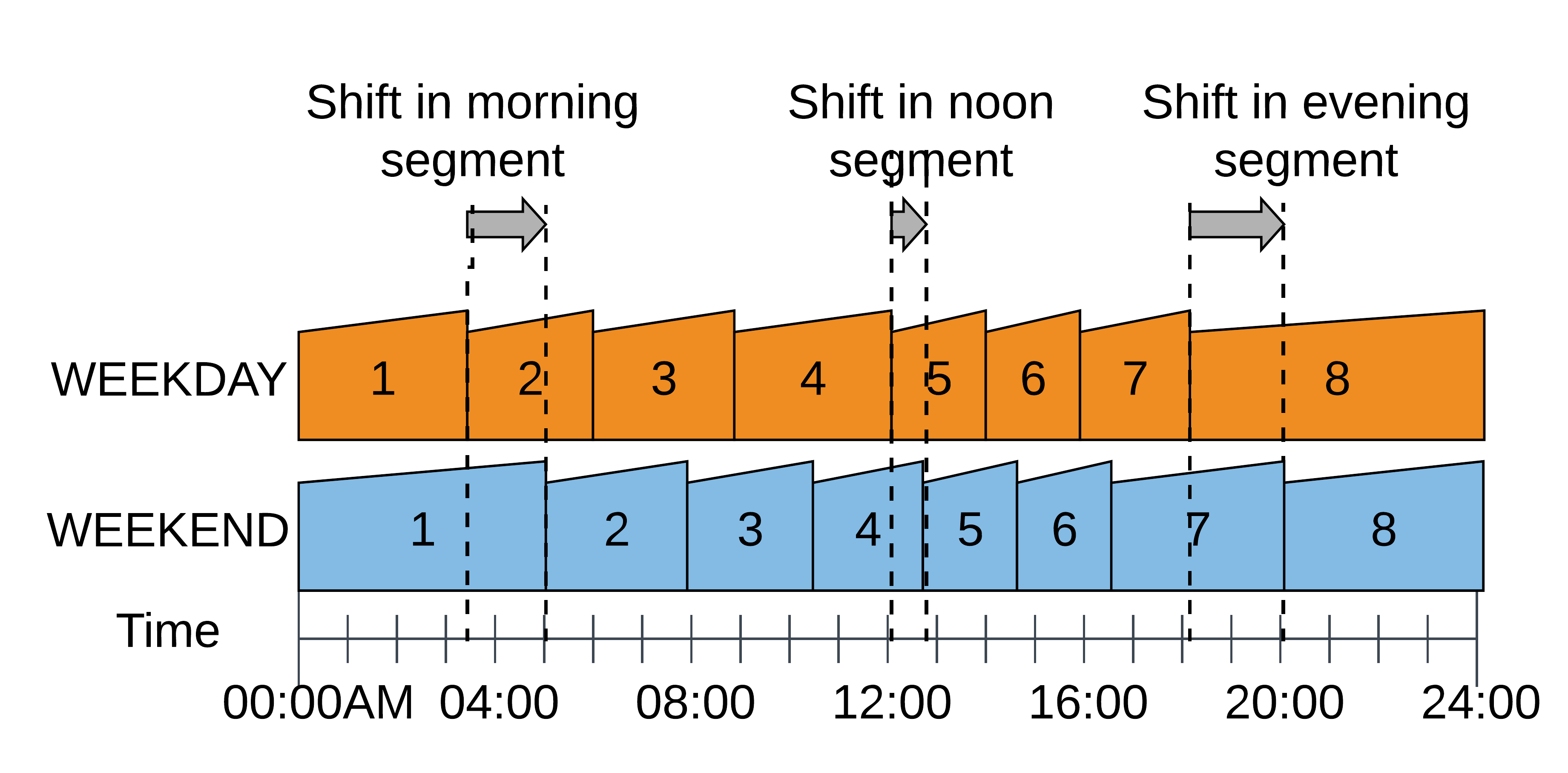}}%
  \caption{ Inferring daily routine from weekday(orange) and weekends(blue) dividing each day into: a) 4, b) 8 segments.}
  \label{fig:wsdm}
\end{figure}

Figure \ref{fig:wsdm} shows the average extracted transition times over 19 weekdays (orange) and 8 holiday/weekend (blue). As we increase the number of segments in a day, the granularity of the task routine increased. There are several distinct differences
between weekdays and weekends in terms of transition times, segment lengths at different times of the day, and the timing of the first and last estimated segments (morning and evening). For example, during the weekdays, the user usually starts the day at around 4:00 am, whilst at weekends, this segment starts at approximately 5:00 am. It should
be noted that this start time is associated with changes in the biometric parameters and may be related to the subject's biological clock. It also shows that users are likely to wake up later on the weekends due to having no work obligations. There was similar shifts in segment transitions in the middle of the day (noon) and later in the day (evening) which are highlighted using grey arrows in the Figure. 
To detect unusual daily patterns, we devised a threshold-based algorithm to find any deviation from normal daily routines. We analyzed the deviation in daily patterns from the reference routine and consider it as an \textit{unusual day} if the dissimilarity metric was greater than a predefined threshold. Using the corresponding images as ground truth, we attempt to explain the reasons behind the strong deviation in activity levels. Table \ref{tab:wsdm_dev} shows the \textit{unusual day}s that had been identified and provides an interpretation of each day. 

\begin{table}[]
  \caption{List of identified unusual days}
  \label{tab:wsdm_dev}
  \begin{tabular}{cl}
   \hline
    Date&Reason of deviation\\
    \hline
    17-8-2016 & The user left work earlier to shop and have lunch\\
    24-8-2016 & The user did not go to work. \\
    29-8-2016 & The user caught a bus instead of driving\\
    30-8-2016 and 8-9-2016 & The user caught a flight  and then went back to work.\\
  \hline
\end{tabular}
\end{table}

\subsection{Case Study2: Detecting Emotion Changes}
The well-known stress and emotional affect dataset WESAD \cite{wesad2018} were analysed with a new perspective. This dataset includes wearable physiological and motion sensors commonly used in medical applications such as electrocardiogram
(ECG), electromyogram
(EMG), Blood Volume Pulse (BVP), temperature, electrodermal activity (EDA), respiration and the accelerometer. Data is labelled in 5 different categories including: Baseline, Stress, Amusement, Medication and Not Defined (ND).

Through this experiment we aim to answer the following questions: 1) What sort of emotion transitions (for example transition from``Stress'' to ``Meditation'') are more detectable? 2) How long does it take for a physiological response to occur for each category of emotion?

\noindent
\begin{figure}[b]
\centering
  
  \subfigure[][]{%
\label{fig:3emotion}%
  \includegraphics[width=0.49\linewidth]{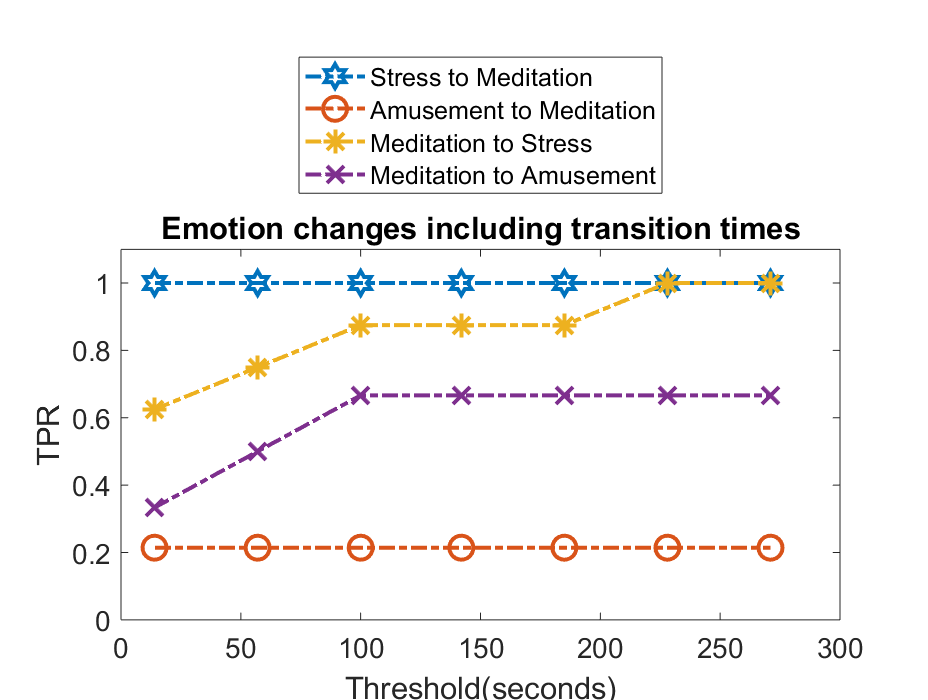}}%
 \hspace{1pt}%
\subfigure[][]{%
\label{fig:emotion_and_transition}%
  \includegraphics[width=0.49\linewidth]{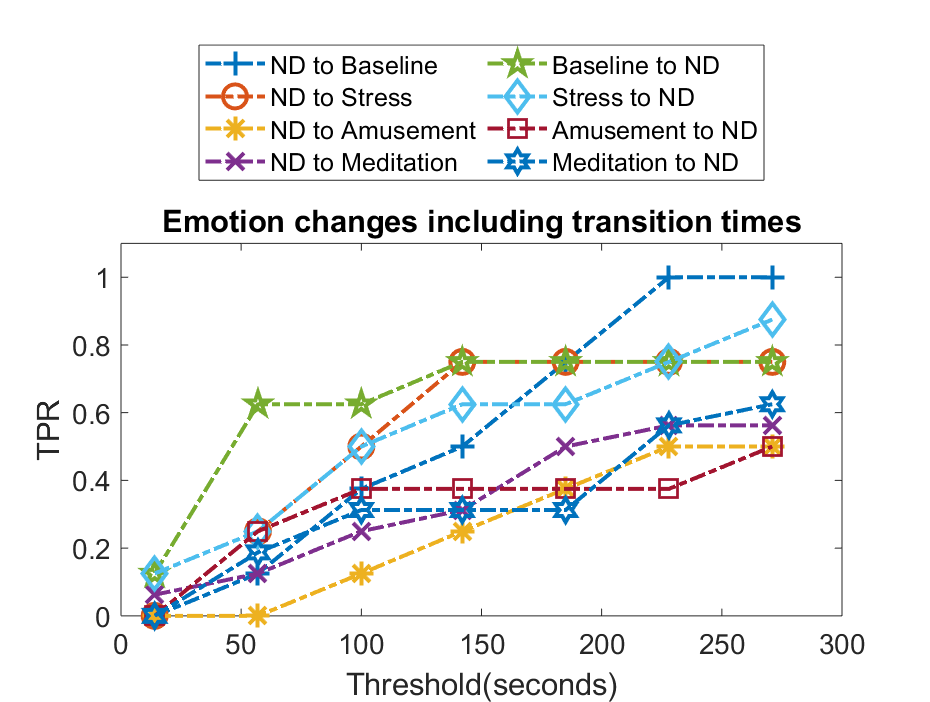}}%
  \caption{The detection of emotion transitions for (a) three emotion states and the (b) whole WESAD dataset}
  \label{fig:wesad_emotions}
\end{figure}

We applied \textit{ESPRESSO} to the data of 15 subjects in two sets of experiments. The first involved estimating the emotion transition times between ``Stress'', ``Amusement'' and ``Meditation'' segments. The second involved estimating emotional transition times across the entire set of data containing Not Defined segments.
For each experiment, we evaluated the estimated emotion transitions in terms of the True Positive Rate (TPR). Section \ref{eval} explains how we consider an extracted boundary as the True Positive. TPR is calculated by dividing the number of true positives by the total number of segments. The detection threshold was changed from 15 seconds to 275 seconds which accounts for the delay in the physiological response.
Figure \ref{fig:wesad_emotions} shows the TPR across different threshold values for the emotion transitions. These figures show that transitions into and out of the ``Stress'' state can be accurately detected in less than 100 seconds. This is due to the strength of the physiological response to ``Stress'' and hints at the negative impact that stress can have on human health. In contrast, we hypothesise the ``Amusement'' to ``Meditation'' transitions are detected with lower accuracy for several reasons. Firstly, the duration of ``Amusement'' segments is much smaller than other emotion states. Secondly, the ``Amusement'' emotion is unlikely to have as strong a physiological effect as ``Stress'' may have. Thirdly, subjectivity is often introduced into such experiments. For the ``Amusement'' segments,  subjects are provided with 11 ``funny'' clips, however, these clips may not be found amusing by all subjects. 
We believe this study opens a new avenue towards improving personality-inference based applications by considering the subject's reactions and response time in different situations dynamically.

\section{Conclusion}
We propose a novel unsupervised method for multivariate time-series segmentation, \textit{ESPRESSO}, and test it on a range of wearable sensing and device-free applications. \textit{ESPRESSO} has a hybrid formulation that enables time series to be segmented on the basis of its temporal shape and statistical properties (i.e. entropy). The proposed temporal shape representation, the Weighted Chained of Arc Curve ($WCAC$), was used to detect potential candidates of segment boundaries. Segments were then estimated with the \textit{GreedyEntropySeg} method that performed a greedy search upon this limited space of boundary candidates using an entropy-based metric.

Our proposed shape segmentation method, Weighted Chained of Arc Curve, $WCAC$, was shown to consistently outperform a state of the art temporal shape based method across three public datasets. This improvement was attributed to our novel primitive, $WCAC$, addressing the limitations of current shape based methods such as segmenting repeated segments or temporal shape patterns that drift over time. 

Experiments were run across a diverse set of seven public datasets of wearable and device free sensors and showed that \textit{ESPRESSO} achieved an average segmentation performance improvement (in terms of F-score and RMSE) over four state of the art methods \textit{FLOSS}, \textit{aHSIC}, \textit{RulSIF} and \textit{IGTS}. Furthermore, 
it was demonstrated that \textit{ESPRESSO} outperformed the four benchmark methods across different categories of the data related to the repetition of patterns and the continuity of segments. An ablation study of \textit{ESPRESSO} demonstrated the $WCAC$ method offered a more significant contribution to segmenting time series with repetitive patterns, whilst the \textit{GreedyEntropySeg} method offered a greater contribution to segmenting time series with non-repetitive patterns and time series composed of non-continuous segments.


We demonstrated the value of using ESPRESSO in two real-world use-cases of inferring daily activity routines and emotional state transitions in an unsupervised context.  

Future work will involve extending the current method to an online version of the channel ranking algorithm where the set of channels accounting for system change can be dynamically selected. Currently, we use the highest ranked channel for segmentation. We also aim to consider subset of channels and their correlation for future segmentation based channel selection.

%

%
 

\begin{acks}
We acknowledge the support of Australian Research Council Discovery DP190101485, and CSIRO Data61 Scholarship program.
\end{acks}

\bibliographystyle{ACM-Reference-Format}
\bibliography{IMWUT-template}

\end{document}